\documentclass[letterpaper,journal]{IEEEtran}
\usepackage{amsmath,amsfonts}
\usepackage{algorithmic}
\usepackage{array}
\usepackage[caption=false,font=normalsize,labelfont=sf,textfont=sf]{subfig}
\usepackage{textcomp}
\usepackage{stfloats}
\usepackage{cuted}
\usepackage[table]{xcolor}
\usepackage{url}
\usepackage{verbatim}
\usepackage{ragged2e}
\usepackage{bm}
\usepackage{tabularx}
\usepackage{multirow}
\usepackage{bbm}
\usepackage{amsmath}
\usepackage{graphicx}
\usepackage{booktabs}
\usepackage{amsbsy}
\usepackage{amssymb}
\usepackage{pifont}
\definecolor{eccvblue}{rgb}{0.21,0.49,0.74}
\definecolor{myorange}{rgb}{0.97,0.69,0.51}
\definecolor{myblue}{rgb}{0,0.75,1}
\definecolor{mypink}{rgb}{1,0,1}
\newcommand{\cmark}{\ding{51}}%
\newcommand{\xmark}{\ding{55}}%
\def\BibTeX{{\rm B\kern-.05em{\sc i\kern-.025em b}\kern-.08em
    T\kern-.1667em\lower.7ex\hbox{E}\kern-.125emX}}
\usepackage{balance}
\usepackage[colorlinks=true, linkcolor=blue, citecolor=blue, urlcolor=blue]{hyperref}
\usepackage[style=numeric-comp, backend=bibtex, sorting=none, minitems=2]{biblatex}
\begin{document}
\title{Diving Deeper into Dense Keypoint Localization in Object Pose Estimation}
\author{
Jiahong Chen, Jinghao Wang, Yueneng Yang, Banglei Guan, \textit{Member, IEEE}, Zi Wang$^*$, Dewen Hu, \textit{Senior Member, IEEE}, and Qifeng Yu
\thanks{This work was supported in part by the Innovation Research Foundation of National University of Defense Technology under Grant 26-ZZCX-JDZ-43 and ZK24-31, the Hunan Provincial Natural Science Foundation of China under Grant 2025JJ60054, and the Hunan Provincial Innovation Foundation for Postgraduate under Grant XJJCB2026108 and XJQY2025054.
(Corresponding authors: Zi Wang.)}
\thanks{Jiahong Chen, Jinghao Wang, Banglei Guan, Zi Wang, and Qifeng Yu are with the College of Aerospace Science and Engineering, National University of Defense Technology, Changsha 410073, China, and also with the Hunan Provincial Key Laboratory of Image Measurement and Vision Navigation, Changsha 410073, China (e-mail: chenjiahong@nudt.edu.cn; wangjinghao16@nudt.edu.cn; guanbanglei12@nudt.edu.cn; wangzi16@nudt.edu.cn; yuqifeng@nudt.edu.cn).
}
\thanks{Yueneng Yang is with the College of Aerospace Science and Engineering, National University of Defense Technology, Changsha 410073, China (e-mail: yangyueneng@163.com).
}
\thanks{Dewen Hu is with the College of Intelligence Science and Technology, National University of Defense Technology, Changsha 410073,
China (e-mail: dwhu@nudt.edu.cn).
}
}

\markboth{IEEE TRANSACTIONS ON CIRCUITS AND SYSTEMS FOR VIDEO TECHNOLOGY}{IEEE TRANSACTIONS ON CIRCUITS AND SYSTEMS FOR VIDEO TECHNOLOGY}

\maketitle

\begin{abstract}
    6D pose estimation of rigid objects is a fundamental and challenging task in computer vision.
    Recent dense semantic keypoint-based methods employ binary encoding for efficient global keypoint localization, delivering strong performance in object pose estimation.
    Nevertheless, the discontinuous nature of binary encoding hinders effective training at high resolutions, while its inherent discrete and quantization errors lead to significant localization uncertainty.
    We propose \textbf{GHLAPose}, a global-to-local framework that synergizes hybrid Gray-binary (HGB) encoding with local-aware refinement (LAR).
    Specifically, HGB encoding first alleviates discontinuity to achieve accurate initial global keypoint localization.
    To further address intrinsic encoding errors, the LAR module estimates uncertainty and refines keypoint localization.
    This synergy enables a precise global-to-local strategy by aggregating global and local features.
    Extensive experiments on LM, LM-O, YCB-V, and TUD-L datasets demonstrate that GHLAPose surpasses state-of-the-art methods, particularly on fine-grained metrics.
    On LM-O, it outperforms the second-best method by 1.4\% and 1.9\% on ADD(-S) 0.02d and 0.05d respectively, and on YCB-V by 1\% on ADD(-S) 0.1d.
    The code will be available at \href{https://github.com/chenjiahongbq/GHLAPose}{\textcolor{magenta}{\textit{GHLAPose}}}.
\end{abstract}


\section{Introduction}
\label{sec:intro}

\IEEEPARstart{O}{bject} 6D pose estimation from a single RGB image has been widely applied in various domains, such as robotic grasping and assembly~\cite{zhong2025dexgrasp,wang2025unigrasptransformer,chen2026learning,MambaX,SpectralGPT}, autonomous driving~\cite{wu20196d,peng2025navigscene}, and augmented reality~\cite{dou2024tactile,peng2026lifelong,hong2026hyperspectral}.
Most existing methods first establish correspondences between 2D image pixels and a set of predefined 3D keypoints~\cite{li2019cdpn,wang2021gdr,tekin2018real,peng2019pvnet,di2021so,lian2023checkerpose,su2022zebrapose,wang2024monocular,zakharov2019dpod,lian2024vapo,park2019pix2pose,OKPOSE,wang2024monocular,DGECN++,wang2025deterministic,Wang_2025_ICCV,chen2026dkpmv}, and then estimate the object pose using the Perspective-n-Point (PnP) solver~\cite{lepetit2009ep,barath2019progressive}. 
This paper follows this two-stage paradigm.

\begin{figure*}[!t]
  \centering
  \refstepcounter{figure}
  \includegraphics[width=0.94\linewidth]{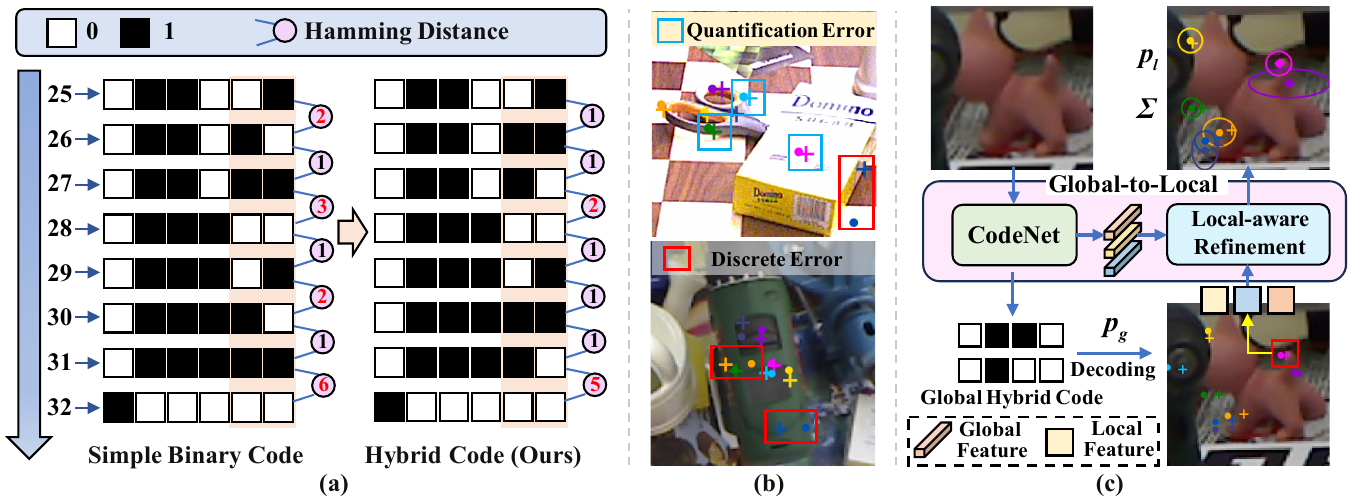}
  \label{fig:motivation}
  \vspace{-3pt}
  \begin{justify}
    \small \textbf{Fig. \arabic{figure}: From encoding challenges to GHLAPose.}
    \textbf{(a)} Simple binary encoding causes spatially adjacent keypoints in the image domain to become distant in the encoding space, especially in the last two bits, as highlighted in the \textcolor{myorange}{orange} region.
    We propose an HGB representation that achieves a more continuous encoding.
    \textbf{(b)} \textcolor{red}{Red} boxes highlight coarse discrete errors in encoding-based localization, while \textcolor{myblue}{blue} boxes denote fine-grained quantization errors.
    Circles denote predicted keypoints, while crosses indicate ground truths.
    \textbf{(c)} GHLAPose first uses CodeNet to generate HGB codes for rapid and accurate global positioning, and then uses LAR for local neighborhood optimization and uncertainty estimation.
  \end{justify}
\end{figure*}

Existing correspondence-based methods can be broadly categorized into two types: estimating the corresponding 3D coordinates on the object surface for each 2D pixel~\cite{li2019cdpn,wang2021gdr,di2021so,su2022zebrapose,zakharov2019dpod,park2019pix2pose,Wang_2025_ICCV} and localizing predefined 3D keypoints in the input image to establish 2D--3D correspondences~\cite{tekin2018real,peng2019pvnet,lian2023checkerpose,wang2024monocular,lian2024vapo,OKPOSE,DGECN++,xu20246d}.
Recent keypoint-based methods~\cite{lian2023checkerpose,lian2024vapo} adopt this formulation and encode dense keypoint locations as compact binary codes.
The predicted codes are decoded into grid coordinates, enabling efficient global localization.
However, simple binary encoding leads to large distances between adjacent keypoints in the encoding space, as shown in Fig.~\ref{fig:motivation}a.
We refer to this as \textit{large transitions}.
A small spatial displacement across a binary carry boundary can change several target bits simultaneously, requiring the network to distinguish substantially different labels from similar image evidence.
This mismatch makes fine-grained localization difficult to learn, particularly for the frequently changing low-order bits.
Discrete coordinate representation also limits localization precision.
Even a correctly predicted code maps to a grid-cell center and leaves a quantization residual relative to the continuous keypoint projection.
Incorrect code predictions additionally assign keypoints to wrong cells, producing discrete localization errors (Fig.~\ref{fig:motivation}b).
Increasing the code length reduces the grid spacing but demands finer spatial discrimination, making code prediction more difficult to learn. Moreover, improving code classification alone cannot recover continuous positions within a cell.
Accurate localization therefore requires both a more suitable encoding representation and correction of the residual coordinate errors after decoding.

Existing works have sought to alleviate these limitations from different perspectives.
For instance, VAPO~\cite{lian2024vapo} introduces visibility-based keypoint filtering to mitigate unreliable predictions, yielding notable performance gains.
Meanwhile, several approaches~\cite{peng2019pvnet,EPro-PnP,Liu_2023_ICCV,moon2025co,wang2024monocular,wang2025deterministic,DGECN++} estimate the uncertainty of keypoints or correspondences and incorporate it into pose optimization.
These strategies improve correspondence selection or weighting, while leaving the fundamental issues of \textit{large transitions} and inherent quantization errors in binary coordinate representations unresolved.

\begin{figure}[!t]
  \centering
  \refstepcounter{figure}
  \includegraphics[width=0.97\linewidth]{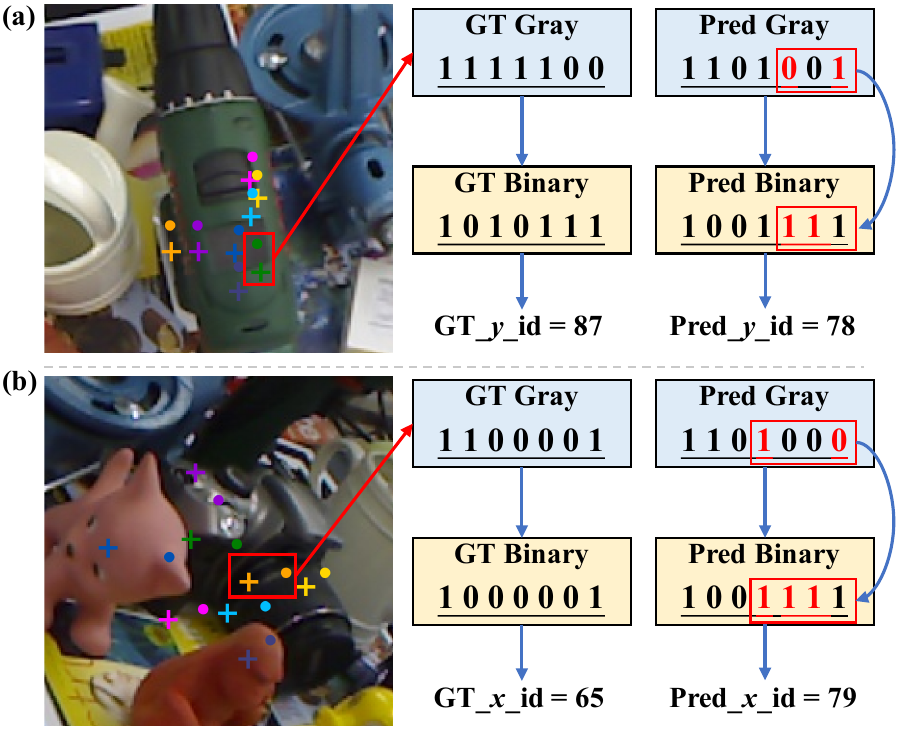}
  \label{fig:gray_case}
  \vspace{-3pt}
  \begin{justify}
    \small \textbf{Fig. \arabic{figure}: Case analysis of error amplification caused by simple Gray code decoding.}
    \textbf{(a)} Error propagation from lower to higher bits.
    \textbf{(b)} Error propagation from few bits to multiple bits.
    Both cases can cause significant discrete errors in keypoint localization.
  \end{justify}
\end{figure}

To bridge this gap, we propose \textbf{GHLAPose}, a global-to-local dense keypoint localization framework illustrated in Fig.~\ref{fig:motivation}c.
The framework first introduces a hybrid Gray-binary (HGB) representation (Fig.~\ref{fig:motivation}a), where Gray code is selectively applied to the lower bits to exploit its spatial continuity for local smoothness.
Since high-order Gray code transitions are susceptible to the significant decoding deviations illustrated in Fig.~\ref{fig:gray_case}, we retain binary encoding for higher bits to ensure robust and rapid global localization.
This hybrid design achieves accurate initial global keypoint localization, providing a robust foundation for the subsequent refinement module.
To further address the residual errors that remain after discrete decoding, we introduce local-aware refinement (LAR), which transitions keypoint localization from the discrete code space to continuous residual estimation. Specifically, LAR anchors local feature sampling at the decoded coordinates, extracting fine-grained image evidence from their neighborhoods and integrating it with global geometry-aware graph features. 
Based on this representation, LAR predicts a continuous sub-pixel displacement to correct residual discrete and quantization errors, while simultaneously estimating keypoint localization uncertainty. 
The estimated uncertainty is further incorporated into nonlinear pose refinement to account for the reliability of individual correspondences.
We conduct extensive experiments on several benchmarks, 
including LM~\cite{LM}, LM-O~\cite{LMO}, YCB-V~\cite{YCB-V}, and TUD-L~\cite{hodan2018bop}. 
GHLAPose consistently outperforms existing state-of-the-art methods, 
particularly on fine-grained metrics such as
ADD(-S) 0.02d and 0.05d.

To summarize, our main contributions are as follows:
\begin{itemize}
    \item We propose GHLAPose, a global-to-local dense keypoint localization framework.
    It first performs encoding-based global localization and then conducts local refinement, enabling accurate and efficient 2D--3D correspondence estimation.

    \item We design HGB to address the \textit{large transitions} inherent in simple binary encoding, which effectively reduces training difficulty and ensures accurate global localization.
    This serves as a robust foundation for subsequent local refinement.
    
    \item We propose LAR, which aggregates global and local features to perform sub-pixel refinement and estimate keypoint uncertainty. 
    LAR effectively mitigates the discrete bias and intrinsic quantization errors inherent in the encoding scheme, complementing the global encoding to establish a robust global-to-local localization framework.
\end{itemize}

The remainder of this paper is organized as follows.
Sec.~\ref{sec:RelatedWork} reviews related work on correspondence-based pose estimation, keypoint representations, and uncertainty estimation.
Sec.~\ref{sec:method} presents the proposed GHLAPose framework.
Sec.~\ref{sec:experiment} reports experimental results and analysis, and Sec.~\ref{sec:Conclusion} concludes this paper.

\section{Related Work}
\label{sec:RelatedWork}

\subsection{Correspondence-Based Pose Estimation}
Most RGB-based 6D pose estimation methods first establish 2D--3D correspondences and then recover the object pose with a PnP solver~\cite{lepetit2009ep,barath2019progressive}.
Early keypoint-based methods construct sparse correspondences.
BB8~\cite{rad2017bb8} and Tekin et al.~\cite{tekin2018real} regress the projections of 3D bounding box corners.
Oberweger et al.~\cite{oberweger2018making} localize such points with heatmaps designed for partial occlusion.
PVNet~\cite{peng2019pvnet} improves robustness by predicting pixel-wise vectors that vote for FPS-sampled keypoints.
Hu et al.~\cite{hu2019segmentation,hu2021wide} further extend sparse keypoint pose estimation with segmentation cues and wider depth ranges.
Although sparse keypoints are efficient, they provide limited geometric constraints when many points are occluded.
Dense correspondence methods address this limitation by predicting object surface coordinates for visible pixels.
CDPN~\cite{li2019cdpn} disentangles coordinate prediction from pose regression.
Pix2Pose~\cite{park2019pix2pose} predicts pixel-level object coordinates with an autoencoder, while GDR-Net~\cite{wang2021gdr} uses geometry-guided dense regression to improve monocular pose estimation.
To make dense prediction more structured, DPOD~\cite{zakharov2019dpod} discretizes dense object coordinates, and ZebraPose~\cite{su2022zebrapose} encodes surface locations in a coarse-to-fine manner.
HccePose~\cite{Wang_2025_ICCV} constructs ultra-dense correspondences from front and back surfaces.
These pixel-wise dense correspondences have achieved strong performance, but the large number of correspondences can make PnP solving time-consuming.
Our method establishes correspondences by globally localizing a compact set of predefined keypoints and refining their image coordinates locally.
This design concentrates correspondence prediction on keypoint locations and reduces the number of correspondences involved in pose solving compared with dense surface mappings.

\subsection{Keypoint and Coordinate Representations}
Keypoint-based pose estimation has gradually moved from sparse regression to dense localization.
Direct regression methods predict keypoint coordinates from image features~\cite{rad2017bb8,park2019pix2pose,zhang2021keypoint}, while heatmap-based methods represent keypoints as spatial response maps~\cite{oberweger2018making}.
Voting methods such as PVNet~\cite{peng2019pvnet} and PVN3D~\cite{he2020pvn3d} aggregate pixel predictions to infer keypoints, which improves robustness but still relies on a relatively small keypoint set.
CheckerPose~\cite{lian2023checkerpose} takes a further step by localizing dense predefined keypoints through simple binary codes.
This formulation turns global localization into a set of compact classification tasks.
VAPO~\cite{lian2024vapo} builds on this dense keypoint formulation and uses visibility-aware keypoint selection to reduce unreliable correspondences.
These methods demonstrate the effectiveness of simple binary codes for dense keypoint representation.
However, their \textit{large transitions} still increase training difficulty and affect global localization accuracy, while the inherent discrete and quantization errors remain unresolved.
6D-Diff~\cite{xu20246d} pursues accurate keypoint localization from another direction by starting from heatmaps and refining them with a diffusion process.
However, this global iterative refinement introduces substantial computation, which limits its efficiency in practical pose estimation.
Surface encoding methods such as ZebraPose~\cite{su2022zebrapose} and HccePose~\cite{Wang_2025_ICCV} improve coordinate representation for dense surface correspondences, but their designs do not directly address dense 2D keypoint localization.
Our HGB representation retains high-order binary bits for coarse localization and applies Gray encoding to the lowest two bits to smooth local code transitions.
This adapts the binary representation used in dense keypoint localization to the different roles of coarse position and fine spatial variation.

\subsection{Uncertainty Estimation and Local Refinement}
Handling unreliable keypoints or correspondences is crucial for accurate pose estimation, particularly in the context of dense correspondence~\cite{lian2024vapo}.
RANSAC~\cite{fischler1981random} removes outliers during PnP solving, but it does not provide fine-grained reliability for individual correspondences.
PVNet~\cite{peng2019pvnet} estimates keypoint uncertainty from voting distributions and uses it in pose estimation.
EPro-PnP~\cite{EPro-PnP} introduces a probabilistic PnP layer that enables end-to-end pose learning from predicted correspondences and their weights.
LC loss~\cite{Liu_2023_ICCV} avoids directly differentiating through PnP by linearizing pose covariance with respect to correspondence covariance.
Co-op~\cite{moon2025co} further refines dense correspondences and couples their uncertainty with differentiable PnP for novel object pose estimation.
These studies demonstrate the importance of modeling correspondence uncertainty for robust pose estimation.
In dense keypoint localization, VAPO~\cite{lian2024vapo} uses visibility awareness to select reliable keypoints, which improves robustness but discards uncertain correspondences.
6D-Diff~\cite{xu20246d} refines keypoints with a diffusion model, but does not model keypoint uncertainty for pose solving.
Our LAR module fuses local image features with global geometric features to predict continuous coordinate residuals and localization uncertainty.
It corrects the globally decoded keypoints and supplies uncertainty weights for subsequent nonlinear pose refinement, coupling correspondence correction with reliability estimation.

\begin{figure*}[!t]
  \centering
  \refstepcounter{figure}
  \includegraphics[width=\linewidth]{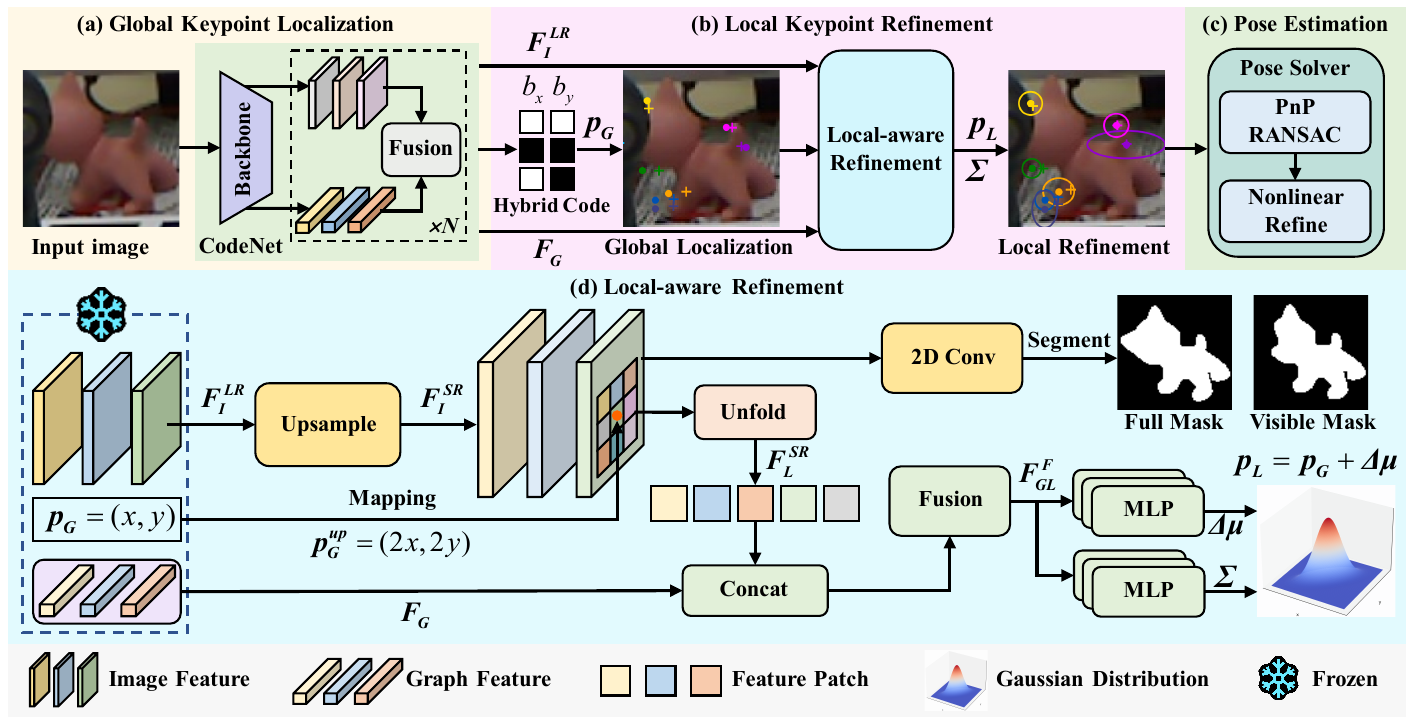}
  \label{fig:pipeline}
  \vspace{-13pt}
  \begin{justify}
    \small \textbf{Fig. \arabic{figure}: Framework of GHLAPose.}
    Given a cropped RGB image, GHLAPose first establishes dense 2D--3D correspondences through global-to-local keypoint localization and then estimates the object pose.
    \textbf{(a)} CodeNet predicts HGB codes for global keypoint localization.
    \textbf{(b)} LAR locally refines keypoints and estimates their uncertainty.
    \textbf{(c)} The pose is solved by PnP/RANSAC~\cite{lepetit2009ep} and further refined with nonlinear optimization.
    \textbf{(d)} LAR fuses local image features with global graph features for local correction and uncertainty prediction.
  \end{justify}
\end{figure*}

\section{Method}
\label{sec:method}
Given an RGB image $\mathcal{I}$ and a rigid object $\mathcal{O}$, our goal is to estimate rotation $\mathbf{R}\in \mathrm{SO}(3)$ and translation $\mathbf{t}\in \mathbb{R}^{3}$ relative to the camera. 
A set of $N$ 3D keypoints $ \boldsymbol{P}\in \mathbb{R}^{3} $ is uniformly sampled from each CAD model using farthest point sampling (FPS) following~\cite{lian2023checkerpose}. 
For the input images, we first employ an off-the-shelf object detector~\cite{ren2015faster, tian2019fcos} to detect the object bounding box and crop its region of interest (RoI), following~\cite{li2019cdpn,su2022zebrapose,wang2021gdr,lian2023checkerpose,lian2024vapo}.

Fig.~\ref{fig:pipeline} illustrates the overall pipeline of GHLAPose.
Given a cropped RGB image, CodeNet first predicts the HGB codes and visibility code $\boldsymbol{b}_{v}$ for dense predefined keypoints.
The predicted HGB codes are decoded to obtain the global keypoint localization $\boldsymbol{p}_{G}$, which provides an efficient coarse 2D--3D correspondence estimate.
LAR then samples local super-resolution features around $\boldsymbol{p}_{G}$ and predicts both the refined keypoint localization $\boldsymbol{p}_{L}$ and its uncertainty $\boldsymbol{\Sigma}$.
During inference, we discard keypoints indicated by $\boldsymbol{b}_{v}$ and use $M_{vis}$ for textureless or self-occluded objects, following~\cite{lian2023checkerpose}.
We first employ PnP/RANSAC~\cite{lepetit2009ep} to estimate the initial object pose and inliers from the dense 2D--3D correspondences.
The inliers are then refined with the Ceres Solver~\cite{Agarwal_Ceres_Solver_2022} through weighted nonlinear optimization, where the predicted keypoint uncertainty $\boldsymbol{\Sigma}$ serves as the weights.

The remainder of this section introduces each component of the pipeline.
Sec.~\ref{sec:HybridRepresentation} presents the HGB representation for efficient and stable global keypoint localization.
Sec.~\ref{sec:Global2local} describes the global-to-local keypoint localization strategy.
Sec.~\ref{sec:Training} details the two-stage training strategy and loss functions.

\subsection{HGB Representation of 2D Keypoints}
\label{sec:HybridRepresentation}
\begin{figure}[!t]
  \centering
  \refstepcounter{figure}
  \includegraphics[width=0.98\linewidth]{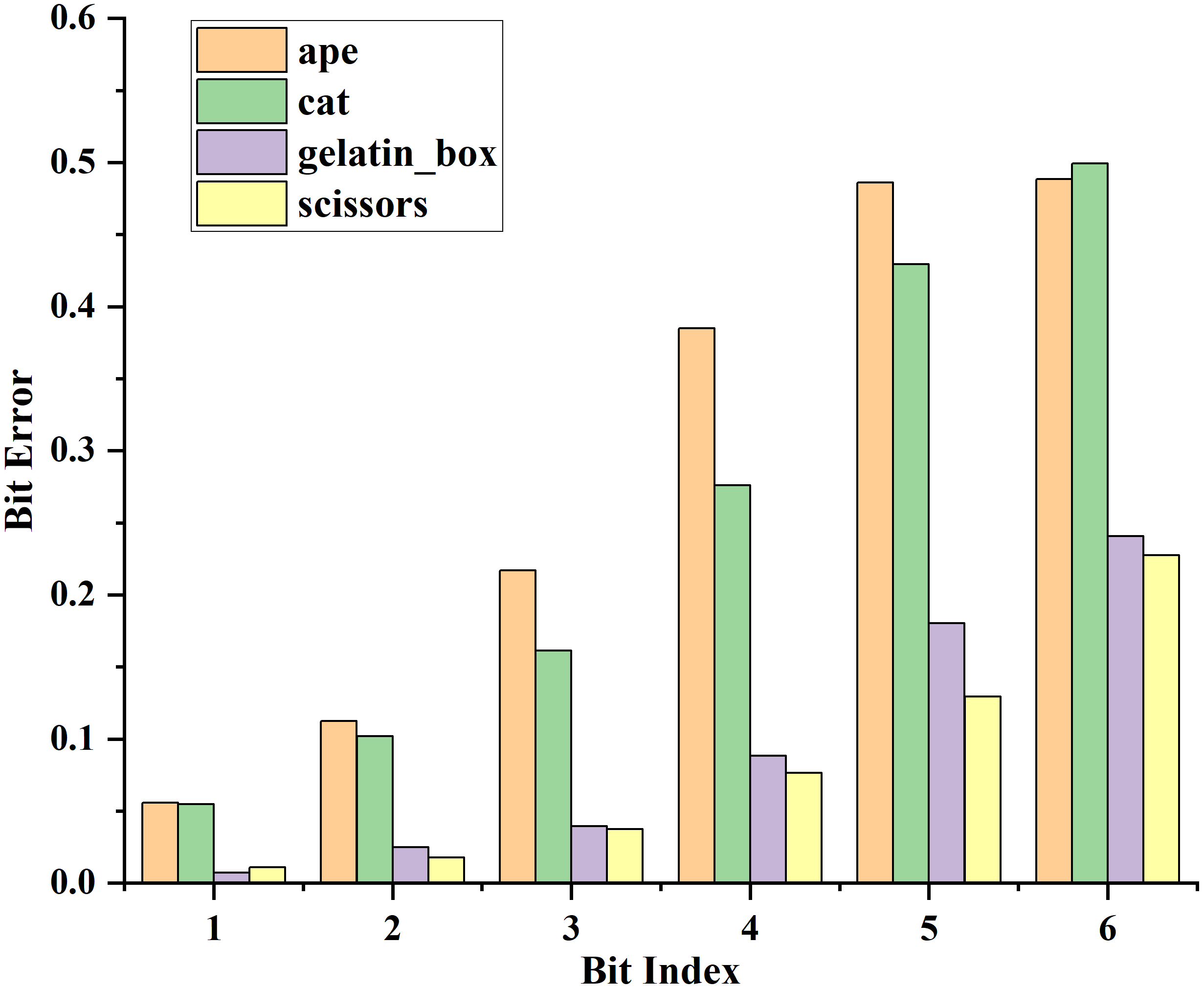}
  \label{fig:binary_err}
  \begin{justify}
    \small \textbf{Fig. \arabic{figure}: Error analysis of previous binary encoding.}
    We analyze the bit-wise errors of binary encoding-based localization and observe that low-order bits are more error-prone, which directly increases fine localization uncertainty.
  \end{justify}
\end{figure}

In this work, we localize a dense set of predefined 3D keypoints $\mathcal{P} \in \mathbb{R}^{N\times3}$ in the image. 
We first represent each 2D keypoint projection with simple binary codes, which have been shown to be efficient for dense keypoint localization~\cite{lian2023checkerpose,lian2024vapo}.
Specifically, the RoI is divided into a $2^d \times 2^d$ grid, and the cell containing the projection $\boldsymbol{p}$ is indexed by $(i_x,i_y)$.
The horizontal index is represented by a $d$-bit binary code $\boldsymbol{b}_{x}$:
\begin{equation}
i_x = \sum_{j=1}^{d} b_x(j) \times 2^{d-j}
\label{eq:binary_code}
\end{equation}
where $b_x(j)$ denotes the $j$-th bit of $\boldsymbol{b}_{x}$, and the vertical index $i_y$ is represented by another $d$-bit code $\boldsymbol{b}_{y}$ in the same way.
The center of the indexed cell is then used as the approximate 2D keypoint location, while an additional visibility code $\boldsymbol{b}_{v}$ indicates whether the keypoint projection lies inside the RoI.

Although this binary representation enables efficient global localization, it suffers from intrinsic encoding defects.
As illustrated in Fig.~\ref{fig:motivation}a, simple binary codes introduce \textit{large transitions} between spatially adjacent locations, especially in the low-order bits.
Fig.~\ref{fig:binary_err} further analyzes binary encoding bit errors and shows that low-order bit errors are closely tied to fine localization uncertainty.
These observations motivate a smoother representation that preserves the efficiency of binary decoding while reducing local discontinuity.

\begin{figure}[!t]
  \centering
  \refstepcounter{figure}
  \includegraphics[width=\linewidth]{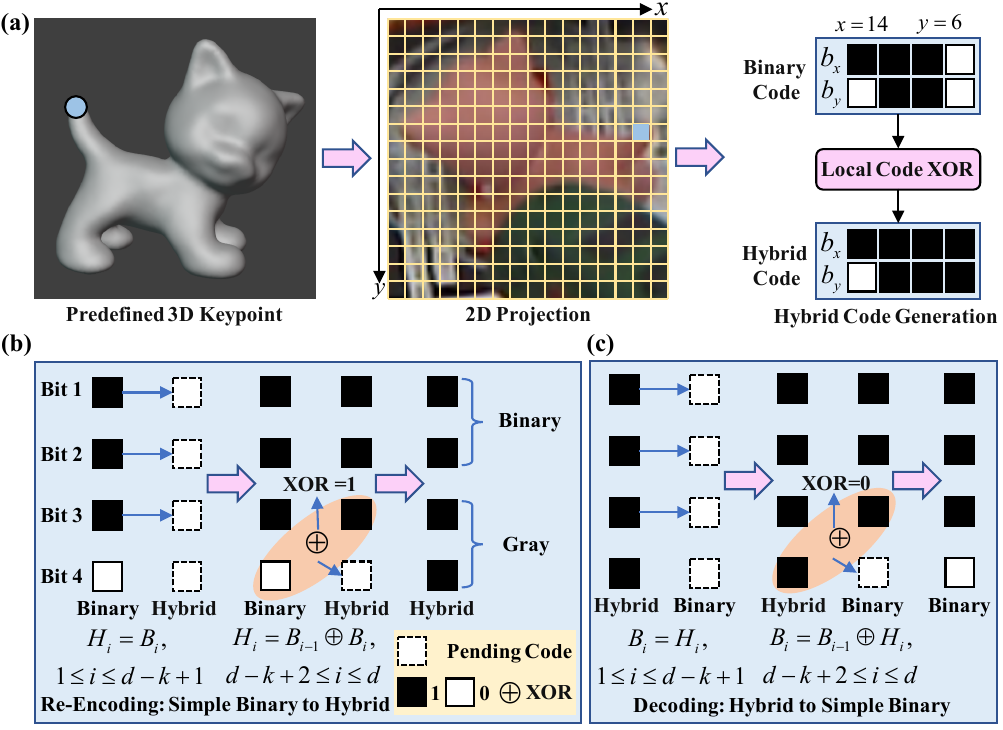}
  \label{fig:hybrid}
  \vspace{-14pt}
  \begin{justify}
    \small \textbf{Fig. \arabic{figure}: Hybrid representation of keypoints.}
    \textbf{(a)} We represent the 2D projections of keypoints using simple binary codes and re-encode these binary codes via local XOR operations to obtain the hybrid encoding.
    \textbf{(b)} The original $d$-bit binary code is transformed into a hybrid encoding, in which the last $k$ bits are Gray codes.
    The \textcolor{myorange}{orange} regions indicate the XOR operation between Gray and binary codes.
    \textbf{(c)} The hybrid encoding is reverted to its simple binary form before mapping it back to coordinates.
  \end{justify}
\end{figure}

To this end, we derive hybrid codes via a local XOR operation, as shown in Fig.~\ref{fig:hybrid}.
Specifically, for a $d$-bit binary code $B$, we convert its last $k$ bits into Gray code to form a hybrid encoding $H$. Since the first bit of a Gray code is identical to that of its binary counterpart, the first $(d-k+1)$ bits remain unchanged:
\begin{equation}
H_{i}=B_{i},\;1\leq i\leq d-k+1
\end{equation}
For the $(d-k+2)$-th to $d$-th bits, we derive the $i$-th Gray code bit by performing an XOR operation between the $(i-1)$-th and $i$-th binary bits:
\begin{equation}
H_{i}=B_{i-1}\oplus B_{i},\;d-k+2\leq i\leq d \label{eq:2}
\end{equation}

When mapping the encoding back to 2D locations, we first decode the hybrid code into its binary form via the inverse of the encoding procedure. The decoded binary codes are then mapped to 2D coordinates according to Eq.~\ref{eq:binary_code}. 
This strategy significantly improves keypoint localization and pose estimation accuracy without additional computational cost, as detailed in Secs.~\ref{sec:AblationStudies} and~\ref{sec:Comparisonsota}.


\textbf{Discussion.}
The proposed hybrid representation is designed to balance local code continuity and decoding robustness. As defined above, HGB preserves the higher-order binary bits while introducing Gray-code transitions only within the low-order block. Since low-order binary bits change more frequently as the coordinate index varies, this selective recoding primarily targets the most frequent local discontinuities. For example, when $k=2$, the two-bit suffix sequence $00,01,10,11$ is transformed into $00,01,11,10$, yielding one-bit transitions within each four-cell block.

For full Gray encoding, the decoding rule and the propagation of prediction errors $e_j=\widehat G_j\oplus G_j$ satisfy:
\begin{equation}
B_i=\bigoplus_{j=1}^{i}G_j,\qquad
\widehat B_i\oplus B_i=\bigoplus_{j=1}^{i}e_j.
\label{eq:gray_error_propagation}
\end{equation}
Thus, a single Gray-bit error at position $j$ affects the decoded binary suffix from $j$ through $d$, and errors at more significant positions can induce larger worst-case coordinate deviations. In contrast, HGB retains direct decoding for the higher-order bits and confines cumulative XOR dependencies to the low-order Gray block. Let $s=d-k+1$ denote the start of this block. When the higher-order prefix is correctly predicted, the coordinate-index error is bounded by:
\begin{equation}
|\widehat x-x|
\leq
\sum_{i=d-k+1}^{d}2^{d-i}
=
2^k-1.
\label{eq:hybrid_error_bound}
\end{equation}
This conditional bound is 3 grid units for $k=2$ and 7 for $k=3$. 
HGB therefore improves local encoding continuity while restricting the spatial extent of coupled decoding errors, providing a principled trade-off between smooth local representation and robust coordinate decoding.

\subsection{Global-to-Local Keypoint Localization}
\label{sec:Global2local}
\textbf{Global Localization.} As illustrated in Fig.~\ref{fig:pipeline}a, we employ an off-the-shelf CodeNet~\cite{lian2023checkerpose} to learn our hybrid encoding. 
Following~\cite{lian2024vapo}, we also incorporate positional encodings for all keypoints.
These encodings are graph features generated by an EdgeConv~\cite{wang2019dynamic} module based on predefined 3D keypoints, effectively embedding spatial relationships into the framework. 
The predicted HGB codes are then decoded back into binary form and subsequently mapped to 2D locations. 
The framework discretizes the search space into a $2^d \times 2^d$ grid $S$ to predict integer coordinates, thereby establishing $\boldsymbol{p}_{G}$ as the \textit{global localization} foundation.

\textbf{Local-aware Refinement.} 
LAR formulates local refinement as continuous residual estimation conditioned on the decoded keypoint location. 
Although HGB improves global localization, incorrect code predictions can still select the wrong grid cell, while even correctly decoded cells leave within-cell positions unresolved. 
Rather than regressing absolute coordinates over the entire RoI, LAR estimates a bounded displacement relative to $\boldsymbol p_G$, using local image evidence and graph-based geometric context. 
This complements discrete global localization by correcting residual cell-selection and quantization errors within a local range.
We integrate local super-resolution features $\boldsymbol{F}_{L}^{SR}$ with global geometry-aware graph features $\boldsymbol{F}_{G}$ to achieve fine-grained localization. 
The architecture of LAR is illustrated in Fig.~\ref{fig:pipeline}d. 
The low-resolution feature map $\boldsymbol{F}_{I}^{LR}$ has half the spatial resolution of the input RoI.
We first upsample it by a factor of 2 to obtain its super-resolution counterpart $\boldsymbol{F}_{I}^{SR}$. 
For each global keypoint localization $\boldsymbol{p}_{G}=(x,y)$, we extract a local super-resolution feature patch $\boldsymbol{F}_{L}^{SR}$ centered at $\boldsymbol{p}_{G}^{up}=(2x,2y)$ with the spatial radius $r$:
\begin{equation}
\begin{aligned}
\Delta_{r}&=\{(i, j) \mid i, j=-r, \ldots, r\} \\
\boldsymbol{F}_{L}^{SR}&=\operatorname{vec}\left([\boldsymbol{F}_{I}^{SR}(2x+i, 2y+j)]_{(i, j) \in \Delta_{r}}\right)
\end{aligned}
\end{equation}
where $\mathrm{vec}(\cdot)$ denotes flattening the tensor into a sequential vector along either rows or columns. 
For the default setting $k=2$, errors confined to the low-order block lead to a maximum coordinate-index deviation of $2^2-1=3$, assuming that the higher-order prefix is correctly predicted. 
With a twofold feature upsampling factor, this corresponds to a sampling radius of $r=2(2^0+2^1)=6$. 
The radius characterizes the spatial scale of low-order decoding errors, while the coordinate correction itself is independently bounded by $R_\mu$. We then fuse $F_L^{SR}$ with $F_G$, where the former provides local image evidence for residual estimation and the latter supplies the geometric context of the corresponding 3D keypoint.

Compared to direct regression, predicting keypoint distributions allows modeling inter-keypoint correlations and quantifying localization uncertainty~\cite{lu2022few}. 
We model the probability density of each refined keypoint using a 2D Gaussian distribution $P_{\Theta}(\boldsymbol{x} \mid \mathcal{I})$, which is formally defined as:
\begin{equation}
    P_{\Theta}(\boldsymbol{x} \mid \mathcal{I})=\frac{1}{2 \pi\left|\boldsymbol{\Sigma}\right|^{1 / 2}} \exp\!\left(-\frac{1}{2}(\boldsymbol{x}-\boldsymbol{\mu}^{\mathrm{gt}})^{\top} \boldsymbol{\Sigma}^{-1}(\boldsymbol{x}-\boldsymbol{\mu}^{\mathrm{gt}})\right)
\end{equation}
where $\Theta$ denotes the model parameters, $\boldsymbol{\Sigma}$ is the covariance matrix specified as $\boldsymbol{\Sigma}=\operatorname{diag}\left(\sigma_{x}^{2}, \sigma_{y}^{2}\right)$, and $\boldsymbol{\mu}^{\mathrm{gt}}=\left[\mu_{x}^{\mathrm{gt}}, \mu_{y}^{\mathrm{gt}}\right]^{\top}$ represents the ground-truth keypoint location.
The LAR predicts the offset $\Delta\boldsymbol{\mu}$ and covariance $\boldsymbol{\Sigma}$ through two separate MLPs:
\begin{equation}
\begin{aligned}
\Delta\boldsymbol{\mu} &= R_{\mu}\tanh\left(\operatorname{MLP}(\boldsymbol{F}_{GL}^{F})\right), \\
\boldsymbol{\Sigma} &= \sigma\left(\operatorname{MLP}(\boldsymbol{F}_{GL}^{F})\right)
\end{aligned}
\end{equation}
$R_{\mu}$ constrains the range of the offset $\Delta\boldsymbol{\mu}$.
Considering the maximum error from the lowest two encoding bits and the twofold feature upsampling, we set $R_{\mu}=7$ to provide a residual correction range suited to these localization errors.
The predicted offset is added to the coarse keypoint $\boldsymbol{p}_{G}$ to obtain the final refined keypoint $\boldsymbol{p}_{L}$:
\begin{equation}
\boldsymbol{p}_{L} = \boldsymbol{p}_{G}+\Delta\boldsymbol{\mu}
\label{eq:refined_keypoint}
\end{equation}

In this stage, we employ local-aware optimization to regress the refined keypoint $\boldsymbol{p}_{L}$ from discrete estimates, achieving continuous sub-pixel accuracy in a process we refer to as \textit{local refinement}.

\subsection{Training}
The training of GHLAPose is conducted in two stages. 
We first train CodeNet independently, after which its parameters are frozen. Subsequently, LAR is further trained on top of the fixed CodeNet.
\label{sec:Training}

\textbf{Training of CodeNet.} CodeNet is trained by supervising the visibility code, the binary index codes, and the auxiliary masks.
Specifically, $\mathcal{L}_{v}$ supervises the visibility code with binary cross-entropy, $\mathcal{L}_{x}$ and $\mathcal{L}_{y}$ supervise the two index codes with bit-wise binary cross-entropy, and $\mathcal{L}_{\text{mask}}$ supervises the full and visible masks with an $L_{1}$ loss.
The CodeNet loss is defined as:
\begin{equation}
    \mathcal{L}_{c}=\mathcal{L}_{v}+\mathcal{L}_{x}+\mathcal{L}_{y}+\mathcal{L}_{\text{mask}}
\end{equation}

\textbf{Training of LAR.} We freeze CodeNet to prevent the discreteness of its predicted coordinates from interfering with the regression loss, while also reducing the overall training complexity. 
Compared to direct keypoint regression using $L_{1}$ or $L_{2}$ loss, we adopt a negative log-likelihood loss (NLL), following~\cite{wang2024monocular,li2021human}.
To obtain the maximum likelihood estimate (MLE) of the normalized variable $\boldsymbol{\mu}$, the NLL loss function for the distribution $P_{\Theta}(\boldsymbol{x} \mid \mathcal{I})$ is defined as:
\begin{equation}
\begin{aligned}
\mathcal{L}_{G}
&=-\left.\log P_{\Theta}(\boldsymbol{x} \mid \mathcal{I})\right|_{\boldsymbol{x}=\boldsymbol{\mu}} \\
&=\log \left(2 \pi\left|\boldsymbol{\Sigma}\right|^{\frac{1}{2}}\right)
+\frac{1}{2}\left(\boldsymbol{\mu}-\boldsymbol{\mu}^{\mathrm{gt}}\right)^{\top}
\boldsymbol{\Sigma}^{-1}\left(\boldsymbol{\mu}-\boldsymbol{\mu}^{\mathrm{gt}}\right)
\end{aligned}
\end{equation}
where $\boldsymbol{\mu}$ denotes the refined keypoint location $\boldsymbol{p}_{L}$ obtained from Eq.~\ref{eq:refined_keypoint}.

We further incorporate a mask $M_{G}$ into the loss term $\mathcal{L}_{G}$ to reduce the sensitivity of the Gaussian distribution to outliers~\cite{yin2025towards}, thereby enhancing training stability.
This design is consistent with the error correction mechanism in \textit{local refinement}, which specifically focuses on addressing global localization errors induced by the last two encoding bits.

Inspired by~\cite{lian2023checkerpose}, we additionally apply a single CNN layer to predict the full segmentation mask $M_{full}$ and the visible mask $M_{vis}$, as illustrated in Fig.~\ref{fig:pipeline}d. We adopt the $L_{1}$ loss as the mask loss $\mathcal{L}_{mask}$, serving as an auxiliary task that explicitly guides the network toward fine-grained super-resolution feature learning.
The overall loss function $\mathcal{L}$ is defined as:
\begin{equation}
    \mathcal{L}=M_{G}\cdot\mathcal{L}_{G}+\lambda\mathcal{L}_{mask}
\end{equation}
where $\lambda$ is a balancing coefficient for the mask loss, used to control the contribution of auxiliary mask supervision relative to the Gaussian localization objective.

\section{Experiments}
\label{sec:experiment}
We evaluate GHLAPose from four complementary aspects.
First, ablation studies analyze the effects of HGB, LAR, and uncertainty-guided pose refinement.
Second, comparisons with state-of-the-art methods on LM, LM-O, YCB-V, and TUD-L validate the overall pose estimation accuracy under standard metrics.
Third, keypoint localization results further examine the quality of the predicted correspondences.
Finally, qualitative and runtime analyses study the localization behavior and practical efficiency of the proposed framework.

\subsection{Experimental Setup}
\textbf{Datasets.} Following the common practice, we conduct extensive experiments on LM~\cite{LM}, LM-O~\cite{LMO}, YCB-V~\cite{YCB-V}, and TUD-L~\cite{hodan2018bop}. LM contains 13 objects, and provides around 1,200 real images for each object with mild occlusions. Following~\cite{brachmann2016uncertainty}, we allocate 15\% of the images for training and evaluate on the remaining set. We also use 1,000 synthetic RGB images for each object during training following~\cite{li2019cdpn,wang2021gdr,di2021so,lian2023checkerpose}. LM-O contains 8 objects from LM, and uses the real images from LM as training images. The test set is composed of 1,214 real images with severe occlusions and clutter. YCB-V is composed
of more than 110,000 real images of 21 objects with severe occlusion and clutter. TUD-L comprises three moving objects captured under diverse lighting conditions and varying degrees of occlusion. For training on LM-O, YCB-V, and TUD-L, we also use the physically based rendered data~\cite{hodavn2020bop} as additional training images, following~\cite{lian2023checkerpose,lian2024vapo,Wang_2025_ICCV,gdrnpp}. 

LM supports controlled analysis of fine-grained localization under relatively mild occlusion, while LM-O tests robustness when occlusion and clutter reduce the available image evidence.
YCB-V broadens the evaluation to a larger collection of object shapes and appearances.
TUD-L further evaluates pose estimation under diverse lighting conditions and varying degrees of occlusion.
Together, these benchmarks evaluate the framework across different objects and imaging conditions, with generalization assessed on held-out images of known objects.

\textbf{Implementation Details.} Our method is implemented in PyTorch~\cite{paszke2019pytorch} and optimized by Adam~\cite{Adam} with a batch size of 32.
We resize each input RoI to $256\times256$ pixels.
We use $N=512$ predefined keypoints and the positional encoding in~\cite{lian2024vapo} to construct CodeNet.
For HGB generation, the code length is set to $d=7$, and the last two bits are represented by Gray code, i.e., $k=2$.
Following~\cite{wang2021gdr,lian2023checkerpose,lian2024vapo}, we use the same data augmentation and precomputed object detections.
Specifically, we use Faster R-CNN~\cite{ren2015faster} for LM, FCOS~\cite{tian2019fcos} for LM-O and YCB-V, and YOLOX~\cite{ge2021yolox} for TUD-L.
Following~\cite{lian2024vapo}, we apply symmetry-aware training~\cite{pitteri2019symmetries} for symmetric objects by normalizing equivalent poses before projecting and encoding the predefined keypoints.
We train a unified pose estimator for all 13 LM objects, while training separate pose estimators for each object on LM-O and YCB-V.

On LM~\cite{LM}, YCB-V~\cite{YCB-V}, and TUD-L~\cite{hodan2018bop}, CodeNet first undergoes a $50\text{k}$-step warm-up at a fixed learning rate of $2\times10^{-4}$.
On LM, CodeNet is then trained for $100\text{k}$ steps with a learning rate of $2\times10^{-4}$, followed by $20\text{k}$ steps with a learning rate of $1\times10^{-4}$.
On YCB-V and TUD-L, CodeNet is trained for $250\text{k}$ steps with a fixed learning rate of $1\times10^{-4}$.
On LM-O~\cite{LMO}, CodeNet is initialized from CheckerPose~\cite{lian2023checkerpose} and trained for $200\text{k}$ steps with a fixed learning rate of $1\times10^{-4}$.
After CodeNet is frozen, LAR is trained for $50\text{k}$ steps on LM and $100\text{k}$ steps on LM-O, YCB-V, and TUD-L with an initial learning rate of $2\times10^{-5}$, which is halved halfway through training, and the loss balancing coefficient is set to $\lambda=1$.

\textbf{Evaluation Metrics.} We adopt the commonly used metric ADD(-S) to evaluate the estimated poses. 
We compute ADD(-S) with a threshold of $x$\%, following~\cite{lian2023checkerpose, lian2024vapo,su2022zebrapose}.
For LM, LM-O, and TUD-L, we select $x\in\lbrace 2,5,10 \rbrace$, following~\cite{lian2024vapo}. 
On YCB-V, we set $x = 10$ and report the AUC (area under the curve) of ADD-S and ADD(-S)~\cite{YCB-V}. 
On LM, we also report the ${n}^{\circ}$, $n$ cm metric following~\cite{lian2023checkerpose,di2021so}.
\begin{table*}[!t]
  \centering
  \refstepcounter{table}
  \begin{justify}
    \small Table \Roman{table}: Ablation on the HGB representation. We compare various encoding lengths and schemes. Hybrid2 denotes that the last two bits use Gray encoding.
  \end{justify}
  \label{table:HGB}
  \setlength{\tabcolsep}{8.0pt}
  \begin{tabular}{ccccccccccc}
    \toprule
    \multirow{2}{*}{Row} & \multirow{2}{*}{Num. of bits} & \multicolumn{3}{c}{Encoding scheme} & \multicolumn{3}{c}{ADD(-S)} & \multirow{2}{*}{$2^{\circ}$2cm} & \multirow{2}{*}{$5^{\circ}$5cm} & \multirow{2}{*}{Mean} \\ \cmidrule(lr){3-5} \cmidrule(lr){6-8}
    & & Binary & Gray & Hybrid2 & 0.02d & 0.05d & 0.1d & & & \\ \midrule
    A0 & \multirow{3}{*}{d=6} & \cmark & \xmark & \xmark & \textbf{48.7} & \textbf{85.6} & 97.2 & 81.1 & 98.9 & 82.3 \\
    A1 & & \xmark & \cmark & \xmark & 46.3 & 84.6 & 97.2 & 77.7 & 98.9 & 80.9 \\
    A2 & & \xmark & \xmark & \cmark & 48.1 & 85.4 & \textbf{97.3} & \textbf{82.4} & \textbf{99.2} & \textbf{82.5} \\ \midrule
    B0 & \multirow{3}{*}{d=7} & \cmark & \xmark & \xmark & 52.3 & 87.0 & 97.5 & 81.8 & 99.0 & 83.5 \\
    B1 & & \xmark & \cmark & \xmark & 53.1 & \textbf{87.8} & \textbf{97.7} & 81.3 & 99.1 & 83.8 \\
    B2 & & \xmark & \xmark & \cmark & \textbf{53.6} & 87.5 & \textbf{97.7} & \textbf{84.4} & \textbf{99.2} & \textbf{84.5} \\ \midrule
    C0 & d=8 & \cmark & \xmark & \xmark & 45.7 & 82.8 & 96.6 & 78.0 & 99.0 & 80.4 \\ \bottomrule
  \end{tabular}
\end{table*}

\subsection{Ablation Study}
\label{sec:AblationStudies}
Unless otherwise specified, we conduct ablation studies on LM~\cite{LM} to validate the effectiveness of each component.
On LM, all ablation variants are trained as a single model for all 13 objects.
Unless otherwise specified, we employ the Progressive-X~\cite{barath2019progressive} solver for pose estimation.
We do not employ $M_{vis}$ for correspondence filtering.

\textbf{Effectiveness of HGB Representation.} Table~\ref{table:HGB} analyzes how the code length and representation scheme affect pose estimation.
The results reveal a trade-off between encoding smoothness and decoding robustness.
Full Gray encoding reduces the code transition between adjacent cells, but high-order bit errors may still cause large coordinate deviations after decoding.
For example, the 6-bit Gray variant decreases the $2^\circ$2cm metric by 3.4\% compared with the binary baseline (A1 $vs.$ A0), consistent with the failure cases in Fig.~\ref{fig:gray_case}.
In contrast, HGB preserves binary coding for high-order bits to maintain stable coarse localization, while using Gray coding only for low-order bits to ease local transitions.
This design consistently improves fine-grained pose accuracy under different code lengths, increasing the $2^\circ$2cm metric from 77.7\% to 82.4\% under 6-bit encoding (A1 $vs.$ A2) and from 81.8\% to 84.4\% under 7-bit encoding (B0 $vs.$ B2).
When the code length increases to 7 bits, directly replacing binary code with Gray code can partially alleviate the discontinuity issue, leading to higher ADD(-S) 0.02d and 0.05d scores (B1 $vs.$ B0).
Nevertheless, the lower $2^\circ$2cm score shows that full Gray encoding is still affected by decoding errors in fine-grained pose estimation.
By combining stable high-order binary bits with smooth low-order Gray bits, HGB achieves the best ADD(-S) 0.02d and $2^\circ$2cm scores in the 7-bit setting, surpassing binary coding by 1.3\% and 2.6\%, respectively.
When the code length increases to 8 bits, performance drops under the same training budget, suggesting that excessively fine discretization increases optimization difficulty and weakens reliable global localization.
We therefore use a 7-bit code and further examine the number of Gray bits below.

\textbf{Effect of the Number of Gray Bits.}
Table~\ref{tab:HGB_k} evaluates the Gray-block length $k$ at a fixed code length of $d=7$ on LM~\cite{LM}. 
Relative to binary encoding, Hybrid2 improves the mean score from 83.5\% to 84.5\% and $2^\circ$2cm accuracy from 81.8\% to 84.4\%. 
A larger Gray block extends one-bit adjacency over a wider coordinate range, but also involves more significant bits in the cumulative decoding dependencies described in Eq.~\eqref{eq:gray_error_propagation}. 
The results are consistent with this trade-off: compared with $k=2$, the configurations with $k=3,4,5$ reduce ADD(-S) 0.02d and $2^\circ$2cm accuracy by 0.5--1.1 and 0.9--2.0 percentage points, respectively. 
Although larger blocks improve some less stringent metrics, $k=2$ achieves the best results on these two fine-grained metrics and the highest mean score among the evaluated configurations. 
As shown in Table~\ref{tab:HGB_k_lmo}, we further evaluate the number of Gray bits on LM-O~\cite{LMO} to examine its effectiveness under occlusion.
Compared with binary encoding ($k=0$), $k=2$ achieves better results on multiple metrics. For example, ADD(-S) 0.05d increases from 44.8\% to 45.6\%.
Increasing $k$ to 3 reduces performance below the binary baseline across all metrics, consistent with the greater sensitivity to decoding errors when occlusion makes code prediction more challenging.

These results across LM~\cite{LM} and LM-O~\cite{LMO} support adopting $k=2$ as the default configuration, balancing local encoding continuity with decoding robustness.

\begin{table}[!ht]
  \centering
  \refstepcounter{table}
  \begin{justify}
    \small Table \Roman{table}: Ablation on the number of Gray bits $k$ in HGB on LM with $d=7$. $k=0$ denotes binary encoding. Best results are highlighted in bold.
  \end{justify}
  \label{tab:HGB_k}
  \setlength{\tabcolsep}{8.5pt}
  \begin{tabular}{ccccccc}
    \toprule
    \multirow{2}{*}{$k$} & \multicolumn{3}{c}{ADD(-S)} & \multirow{2}{*}{$2^{\circ}$2cm} & \multirow{2}{*}{$5^{\circ}$5cm} & \multirow{2}{*}{Mean} \\ \cmidrule(lr){2-4}
    & 0.02d & 0.05d & 0.1d & & & \\ \midrule
    0 & 52.3 & 87.0 & 97.5 & 81.8 & 99.0 & 83.5 \\
    2 & \textbf{53.6} & 87.5 & 97.7 & \textbf{84.4} & \textbf{99.2} & \textbf{84.5} \\
    3 & 53.0 & \textbf{88.1} & 97.8 & 83.5 & \textbf{99.2} & 84.3 \\
    4 & 52.5 & 87.5 & 97.5 & 82.4 & 99.1 & 83.8 \\
    5 & 53.1 & 87.8 & \textbf{98.0} & 83.1 & 99.1 & 84.2 \\ \bottomrule
  \end{tabular}
\end{table}

\begin{table}[!ht]
  \centering
  \refstepcounter{table}
  \begin{justify}
    \small Table \Roman{table}: Ablation on the number of Gray bits $k$ in HGB on LM-O. Best results are highlighted in bold.
  \end{justify}
  \label{tab:HGB_k_lmo}
  \setlength{\tabcolsep}{11.0pt}
  \begin{tabular}{ccccc}
    \toprule
    \multirow{2}{*}{$k$} & \multicolumn{3}{c}{ADD(-S)} & \multirow{2}{*}{Mean} \\ \cmidrule(lr){2-4}
    & 0.02d & 0.05d & 0.1d & \\ \midrule
    0 & \textbf{9.8} & 44.8 & 77.0 & 43.9 \\
    2 & \textbf{9.8} & \textbf{45.6} & \textbf{77.3} & \textbf{44.2} \\
    3 & 9.3 & 44.7 & 76.9 & 43.6 \\ \bottomrule
  \end{tabular}
\end{table}

\textbf{Effect of the Residual Range.}
Table~\ref{tab:lar_residual_range} evaluates the effect of the residual range $R_{\mu}$ on LM.
Reducing $R_{\mu}$ from 7 to 4 decreases ADD(-S) 0.02d from 55.1\% to 54.2\% and $2^\circ$2cm accuracy from 86.0\% to 85.5\%, suggesting that a smaller range can limit the correction of localization errors.
Increasing $R_{\mu}$ to 10 expands the range of predicted offsets, which may make fine-grained residual estimation more difficult, slightly reducing these scores to 54.9\% and 85.8\%, respectively.
Among the evaluated settings, $R_{\mu}=7$ achieves the best fine-grained and mean results, supporting our choice of correction range.

\begin{table}[ht]
  \centering
  \refstepcounter{table}
  \begin{justify}
    \small Table \Roman{table}: Ablation on the residual range $R_{\mu}$ of LAR on LM. Best results are highlighted in bold.
  \end{justify}
  \label{tab:lar_residual_range}
  \setlength{\tabcolsep}{8.5pt}
  \begin{tabular}{ccccccc}
    \toprule
    \multirow{2}{*}{$R_{\mu}$} & \multicolumn{3}{c}{ADD(-S)} & \multirow{2}{*}{$2^{\circ}$2cm} & \multirow{2}{*}{$5^{\circ}$5cm} & \multirow{2}{*}{Mean} \\ \cmidrule(lr){2-4}
    & 0.02d & 0.05d & 0.1d & & & \\ \midrule
    4 & 54.2 & 87.8 & 97.5 & 85.5 & \textbf{99.3} & 84.9 \\
    7 & \textbf{55.1} & \textbf{88.3} & \textbf{97.8} & \textbf{86.0} & \textbf{99.3} & \textbf{85.3} \\
    10 & 54.9 & 87.7 & 97.7 & 85.8 & \textbf{99.3} & 85.1 \\ \bottomrule
  \end{tabular}
\end{table}

\begin{figure}[ht]
  \centering
  \refstepcounter{figure}
  \includegraphics[width=\linewidth]{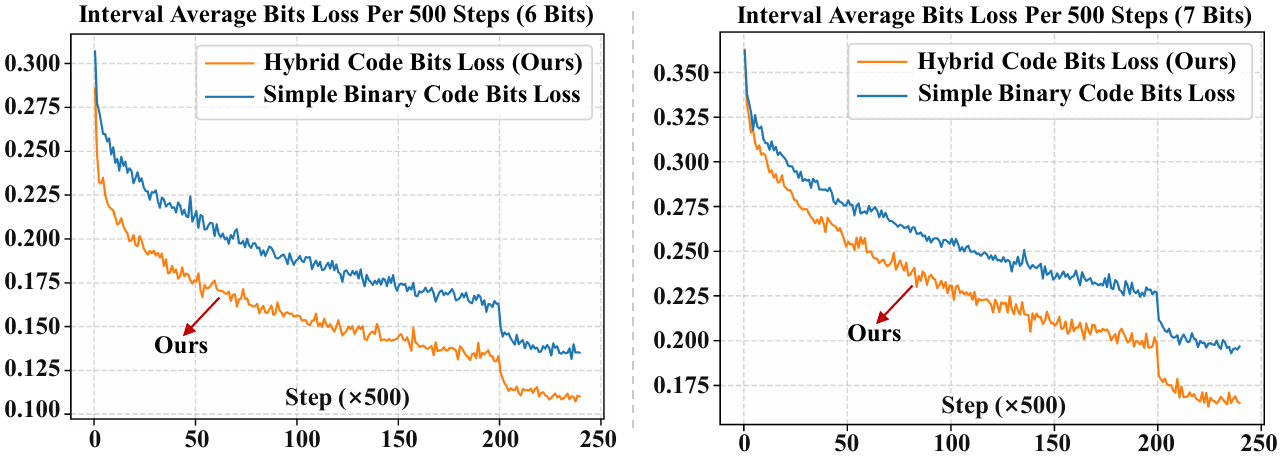}
  \label{fig:train_loss}
  \vspace{-13pt}
  \begin{justify}
    \small \textbf{Fig. \arabic{figure}: Training loss comparison.}
    We show the bit loss for the $x$-coordinate under 6-bit and 7-bit encoding; the trend for the $y$-coordinate is similar.
    HGB reduces the bit encoding loss under both code lengths.
  \end{justify}
\end{figure}

\textbf{Training Stability.} Fig.~\ref{fig:train_loss} further explains the optimization behavior behind Table~\ref{table:HGB}.
Under the same loss setting, HGB produces lower bit losses than binary encoding for both 6-bit and 7-bit settings, indicating that the reduced local transitions make dense code prediction easier to optimize as the encoding resolution increases.

\begin{figure}[ht]
  \centering
  \refstepcounter{figure}
  \includegraphics[width=\linewidth]{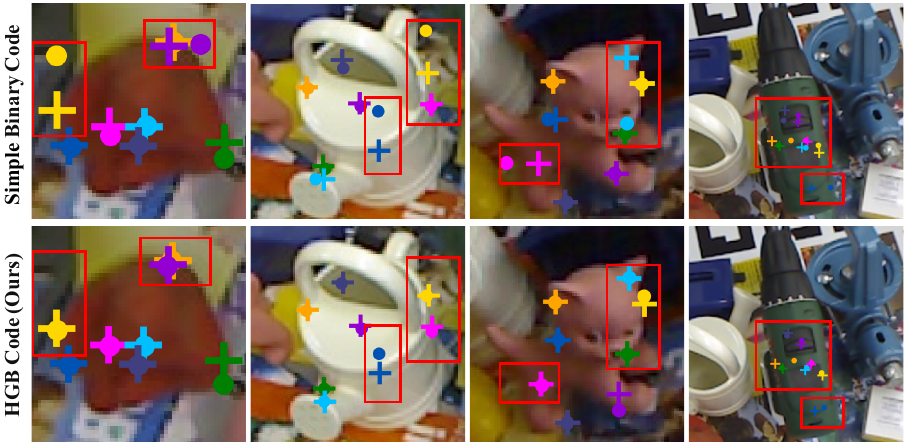}
  \label{fig:localization_HGB}
  \vspace{-13pt}
  \begin{justify}
    \small \textbf{Fig. \arabic{figure}: Qualitative keypoint localization results of the HGB code.}
    \textcolor{red}{Red} boxes highlight that HGB encoding achieves more accurate keypoint localization than simple binary encoding.
  \end{justify}
\end{figure}

\textbf{HGB Keypoint Localization.} Fig.~\ref{fig:localization_HGB} shows the keypoint localization results of HGB.
Compared with simple binary encoding, HGB reduces training difficulty and improves code prediction accuracy, leading to more reliable global keypoint localization.
It also alleviates coarse localization failures caused by high-order bit prediction errors, providing better initial keypoints for the subsequent LAR module.

\textbf{Effectiveness of LAR.} Table~\ref{tab:LAR} evaluates the robustness of LAR and its complementarity with HGB.
Without nonlinear refinement, LAR consistently improves the global localization results under different code lengths and encoding schemes.
For 6-bit HGB, LAR improves ADD(-S) 0.02d from 48.1\% to 52.9\% and the mean score from 82.5\% to 84.2\% (D0 $vs.$ D1).
Under 7-bit encoding, LAR also improves the binary baseline by 1.9\% on ADD(-S) 0.02d (E0 $vs.$ E1) and improves HGB by 1.5\% on ADD(-S) 0.02d and 1.6\% on $2^\circ$2cm (E2 $vs.$ E3).
These consistent gains indicate that LAR robustly compensates for the discrete and quantization errors left by global keypoint localization.

More importantly, LAR works best when paired with HGB.
Without nonlinear refinement, 7-bit HGB with LAR already outperforms 7-bit binary with LAR by 1.0\% on the mean score and 3.0\% on $2^\circ$2cm (E3 $vs.$ E1).
With nonlinear refinement enabled, this advantage remains clear, where 7-bit HGB with LAR achieves 86.2\% mean accuracy and surpasses 7-bit binary with LAR by 1.1\% on average and 2.9\% on $2^\circ$2cm (E5 $vs.$ E4).
This shows that the two modules are complementary: HGB provides more reliable global keypoints, and LAR further corrects their remaining local errors.
Together, they form an effective global-to-local keypoint localization framework.
LAR also incurs low computational overhead, as analyzed in Sec.~\ref{sec:runtime}.

\begin{figure}[!ht]
  \centering
  \refstepcounter{figure}
  \includegraphics[width=\linewidth]{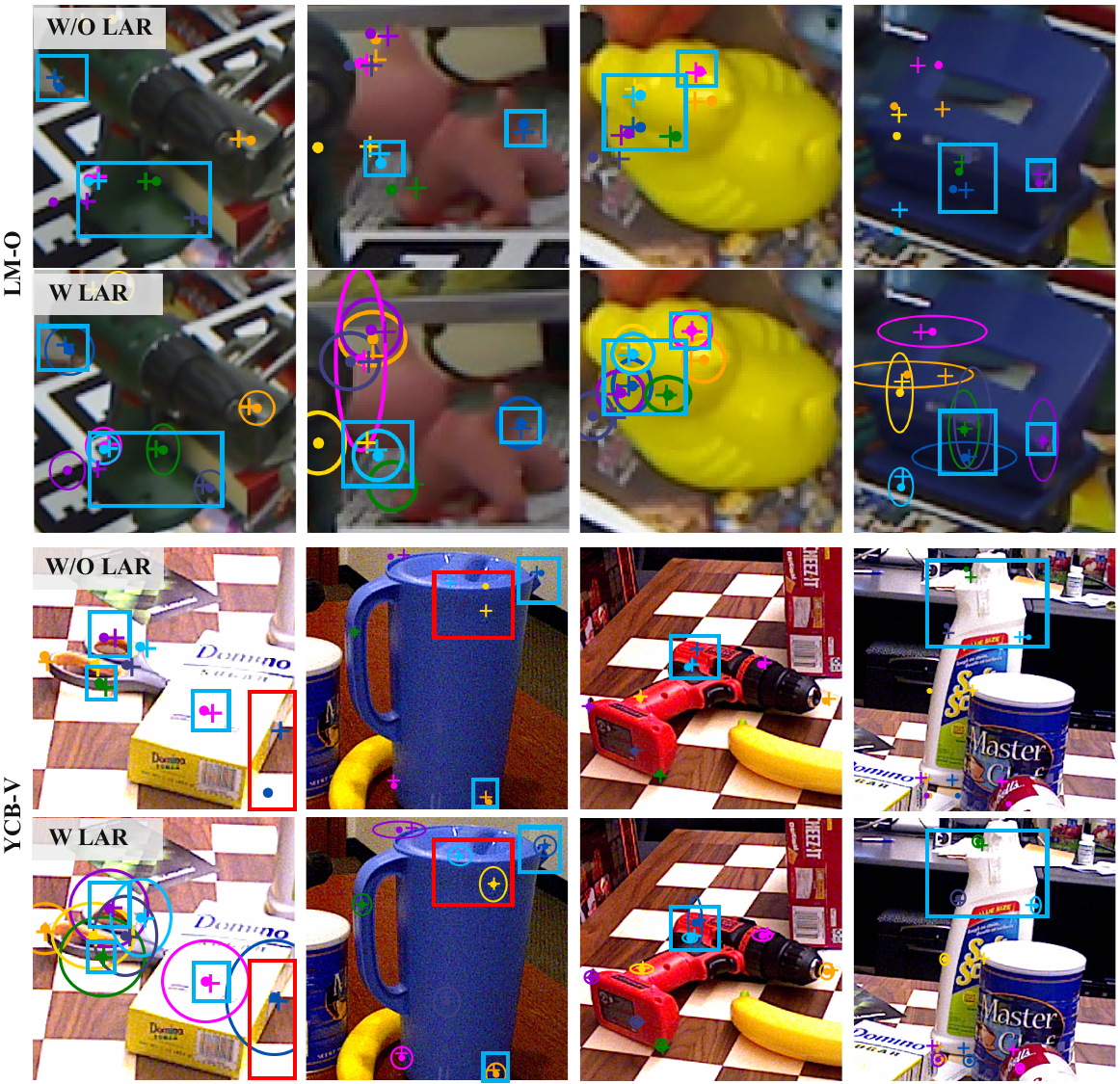}
  \label{fig:localization_LAR}
  \vspace{-13pt}
  \begin{justify}
    \small \textbf{Fig. \arabic{figure}: Qualitative keypoint localization results of LAR.}
    \textcolor{red}{Red} and \textcolor{myblue}{blue} boxes highlight the local refinement of discrete and quantization errors, respectively.
  \end{justify}
\end{figure}

\textbf{LAR Keypoint Localization.} Fig.~\ref{fig:localization_LAR} provides qualitative evidence on the LM-O~\cite{LMO} and YCB-V~\cite{YCB-V} datasets.
LAR refines the coarse global estimates into more accurate coordinates, effectively correcting both large discrete localization errors and local quantization errors highlighted in the visual results.
These refined keypoints provide more reliable 2D--3D correspondences for subsequent pose solving.

\begin{figure}[!t]
  \centering
  \refstepcounter{figure}
  \includegraphics[width=\linewidth,pagebox=cropbox]{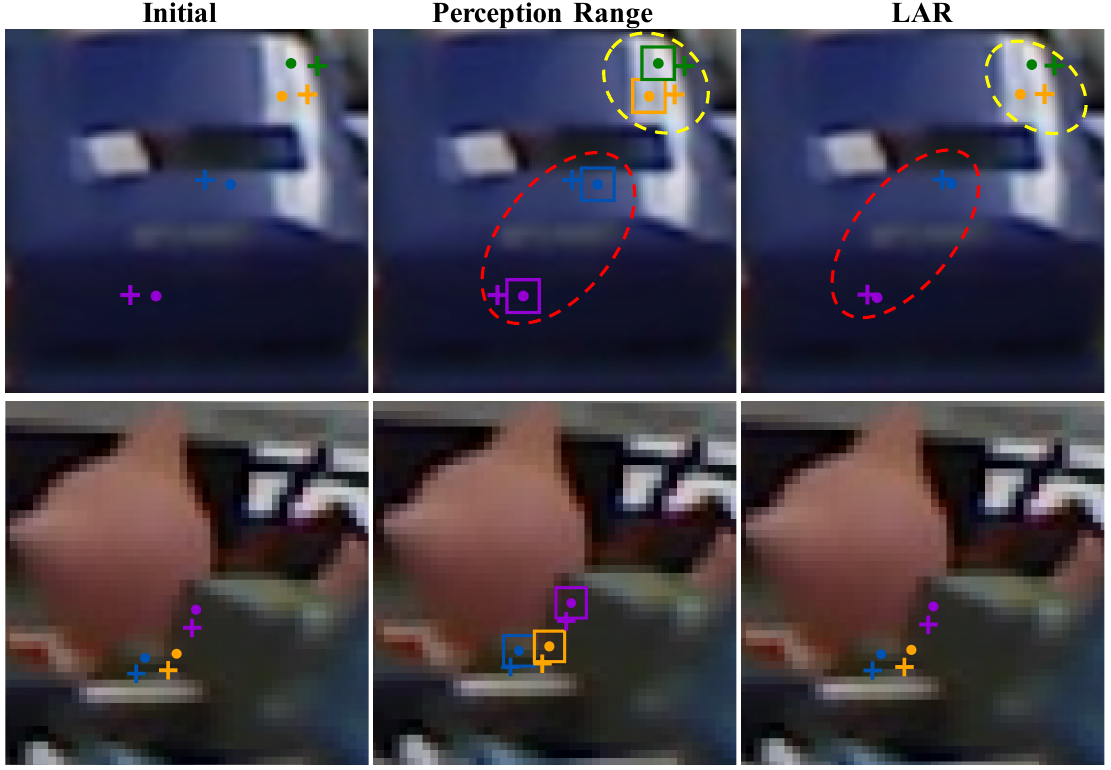}
  \label{fig:lar_behavior}
  \vspace{-5mm}
  \begin{justify}
    \small \textbf{Fig. \arabic{figure}: LAR behavior when ground-truth keypoints lie outside the local sampling range.}
    The top and bottom rows show non-occluded and occluded cases, respectively. Dots denote predicted keypoints, crosses denote ground truths, and squares indicate local sampling regions. Red and yellow ellipses highlight successful corrections and reflection-related failures, respectively.
  \end{justify}
\end{figure}

\textbf{LAR Behavior Analysis.} Fig.~\ref{fig:lar_behavior} examines cases where the ground-truth keypoint lies outside the local feature patch centered at the initial prediction.
In the non-occluded examples highlighted by the red ellipses, LAR moves the predictions toward the ground truth even when the target is outside the sampled region.
This behavior is consistent with its fusion of local image features and global geometric context, which provides information beyond the local patch for predicting correction directions.
The yellow ellipses highlight failure cases where local reflections weaken discriminative image features, leaving the predictions close to their initial positions.
In the occluded examples, the target's local appearance is completely obscured and the sampled features are dominated by the occluder, leaving little useful local evidence for refinement.
The corrections are consequently small, and residual localization errors remain despite the global geometric context.
These cases illustrate the behavior of LAR under different visibility and appearance conditions and support our design of combining global geometric context with local image features for keypoint refinement.

\begin{table*}[!t]
  \centering
  \refstepcounter{table}
  \begin{justify}
    \small Table \Roman{table}: Ablation on LAR. Results demonstrate the robustness of LAR across various encoding bit-widths and schemes. Hybrid2 denotes that the last two bits use Gray encoding. NR denotes nonlinear refinement. (*) denotes the use of the PnP/RANSAC~\cite{lepetit2009ep} solver.
  \end{justify}
  \label{tab:LAR}
  \setlength{\tabcolsep}{7.3pt}
  \begin{tabular}{ccccccccccccc}
    \toprule
    \multirow{2}{*}{Row} & \multirow{2}{*}{Num. of bits} & \multicolumn{2}{c}{Encoding scheme} & \multirow{2}{*}{LAR} & \multirow{2}{*}{NR} & \multirow{2}{*}{$\Sigma$} & \multicolumn{3}{c}{ADD(-S)} & \multirow{2}{*}{$2^{\circ}$2cm} & \multirow{2}{*}{$5^{\circ}$5cm} & \multirow{2}{*}{Mean} \\ \cmidrule(lr){3-4} \cmidrule(lr){8-10}
    & & Binary & Hybrid2 & & & & 0.02d & 0.05d & 0.1d & & & \\ \midrule
    D0 & \multirow{3}{*}{d=6} & \xmark & \cmark & \xmark & \xmark & \xmark & 48.1 & 85.4 & 97.3 & 82.4 & \textbf{99.2} & 82.5 \\
    D1 & & \xmark & \cmark & \cmark & \xmark & \cmark & 52.9 & 87.4 & 97.6 & 84.1 & 99.2 & 84.2 \\
    D2 & & \xmark & \cmark & \cmark & \cmark & \cmark & \textbf{54.3} & \textbf{88.3} & \textbf{97.8} & \textbf{84.5} & \textbf{99.3} & \textbf{84.8} \\ \midrule
    E0 & \multirow{8}{*}{d=7} & \cmark & \xmark & \xmark & \xmark & \xmark & 52.3 & 87.0 & 97.5 & 81.8 & 99.0 & 83.5 \\
    E1 & & \cmark & \xmark & \cmark & \xmark & \cmark & 54.2 & 87.8 & 97.6 & 83.0 & 99.0 & 84.3 \\
    E2 & & \xmark & \cmark & \xmark & \xmark & \xmark & 53.6 & 87.5 & 97.7 & 84.4 & 99.2 & 84.5 \\
    E3 & & \xmark & \cmark & \cmark & \xmark & \cmark & 55.1 & 88.3 & 97.8 & 86.0 & 99.3 & 85.3 \\
    E4 & & \cmark & \xmark & \cmark & \cmark & \cmark & 55.7 & 88.9 & 98.0 & 84.0 & 99.1 & 85.1 \\
    E5 & & \xmark & \cmark & \cmark & \cmark & \cmark & 56.8 & \textbf{89.5} & \textbf{98.3} & \textbf{86.9} & \textbf{99.4} & \textbf{86.2} \\
    E6 & & \xmark & \cmark & \cmark & \cmark & \cmark & \textbf{56.9*} & \textbf{89.5*} & \textbf{98.3*} & 86.8* & \textbf{99.4*} & \textbf{86.2*} \\
    E7 & & \xmark & \cmark & \cmark & \cmark & \xmark & 56.8* & 89.1* & 98.2* & 86.3* & \textbf{99.4*} & 86.0* \\ \bottomrule
  \end{tabular}
\end{table*}

\textbf{Effectiveness of Nonlinear Refinement \& Uncertainty Estimation.} We learn the uncertainty of keypoints and convert it into confidence weights to perform a nonlinear refinement on the initial pose estimates.
As shown in Table~\ref{tab:LAR}, nonlinear refinement consistently boosts pose estimation performance across various encoding bit depths and schemes, particularly in fine-grained metrics.
For instance, with 7-bit HGB, nonlinear refinement yields gains of 1.7\% and 1.2\% in ADD(-S) 0.02d and 0.05d (E5 $vs.$ E3), confirming its ability to enforce global geometric consistency across dense keypoints for sub-pixel alignment.
Crucially, this dense geometric alignment neutralizes initial pose deviations introduced by specific solvers.
By enforcing global consistency, PnP/RANSAC~\cite{lepetit2009ep} achieves performance parity with the typically superior Progressive-X~\cite{barath2019progressive} (E6 $vs.$ E5), demonstrating that our refinement stage effectively mitigates the dependency on initial solver selection.
Consequently, we adopt the simpler PnP/RANSAC~\cite{lepetit2009ep} for initial pose estimation to prioritize runtime efficiency, as discussed in Sec.~\ref{sec:runtime}.
Furthermore, excluding uncertainty estimation leads to a performance decline, particularly on stringent metrics such as $2^\circ$2cm and ADD(-S) 0.05d (E6 $vs.$ E7).
This underscores that uncertainty serves as a critical weighting mechanism, guiding the nonlinear refinement toward more robust and accurate pose optimization.

\textbf{Keypoint Regression vs. Encoding.} Table~\ref{tab:LAR_extra} further compares our encoding-based localization with direct keypoint regression under the same network architecture.
Direct regression leads to a severe performance drop, decreasing the mean score from 84.5\% to 46.2\% without LAR (B2 $vs.$ F0) and from 85.3\% to 47.7\% with LAR (E3 $vs.$ F1).
The fine-grained $2^\circ$2cm metric also drops from 84.4\% to 18.3\% and from 86.0\% to 20.0\%, respectively.
Nevertheless, LAR still improves the regression-based representation, increasing the mean score from 46.2\% to 47.7\% and the $2^\circ$2cm metric from 18.3\% to 20.0\% (F0 $vs.$ F1).
These results indicate that dense keypoint localization benefits substantially from the structured encoding formulation, while direct coordinate regression is difficult to optimize for this task.

\begin{table}[!t]
  \centering
  \refstepcounter{table}
  \begin{justify}
    \small Table \Roman{table}: Additional ablation on regression and loss weighting on the LM~\cite{LM} dataset. Regression denotes directly regressing keypoints under the same network architecture.
  \end{justify}
  \label{tab:LAR_extra}
  \setlength{\tabcolsep}{3.5pt}
  \begin{tabular}{@{}cccccccc@{}}
    \toprule
    \multirow{2}{*}{Row} & \multirow{2}{*}{Method} & \multicolumn{3}{c}{ADD(-S)} & \multirow{2}{*}{$2^{\circ}$2cm} & \multirow{2}{*}{$5^{\circ}$5cm} & \multirow{2}{*}{Mean} \\ \cmidrule(lr){3-5}
    & & 0.02d & 0.05d & 0.1d & & & \\ \midrule
    F0 & B2 $\rightarrow$ w/ Regression & 12.8 & 49.0 & 80.5 & 18.3 & 70.3 & 46.2 \\
    F1 & E3 $\rightarrow$ w/ Regression & 14.8 & 50.7 & 81.2 & 20.0 & 71.7 & 47.7 \\ \midrule
    G0 & E3 $\rightarrow$ $\lambda=0.3$ & 55.3 & 88.0 & 97.7 & 85.7 & 99.3 & 85.2 \\
    G1 & E3 $\rightarrow$ $\lambda=0.5$ & 54.7 & 88.0 & 97.8 & 85.6 & 99.3 & 85.1 \\ \bottomrule
  \end{tabular}
\end{table}

\textbf{Sensitivity to Loss Weight.} We also study the influence of the mask loss weight $\lambda$ in Table~\ref{tab:LAR_extra}.
Compared with the default setting $\lambda=1$ (E3), using $\lambda=0.3$ and $\lambda=0.5$ produces similar mean scores of 85.2\% and 85.1\%, respectively.
This suggests that LAR is not sensitive to moderate changes of $\lambda$.
Since $\lambda=1$ achieves the highest mean accuracy, we use it in all main experiments.

\begin{table}[ht]
  \centering
  \refstepcounter{table}
  \begin{justify}
    \small Table \Roman{table}: Comparison with SOTA methods on the LM~\cite{LM} dataset. We highlight the best results in \textbf{bold} and \underline{underline} the second-best results. ``--'' denotes unavailable results.
  \end{justify}
  \label{table:Comparison_lm}
  \setlength{\tabcolsep}{5.5pt}
  \begin{tabular}{lcccccc}
    \toprule
    \multirow{2}{*}{Method} & \multicolumn{3}{c}{ADD(-S)} & \multirow{2}{*}{$2^{\circ}$2cm} & \multirow{2}{*}{$5^{\circ}$5cm} & \multirow{2}{*}{Mean} \\ \cmidrule(lr){2-4}
    & 0.02d & 0.05d & 0.1d & & & \\ \midrule
    GDR-Net~\cite{wang2021gdr} & 35.5 & 76.3 & 93.7 & 62.1 & -- & -- \\
    NVR-Net~\cite{NVR-Net} & 41.7 & 79.8 & 95.3 & 74.6 & -- & -- \\
    SO-Pose~\cite{di2021so} & 45.9 & 83.1 & 96.0 & 76.9 & 98.5 & 80.1 \\
    EPro-PnP~\cite{EPro-PnP} & 44.8 & 82.0 & 95.8 & 81.0 & 98.5 & 80.4 \\
    CheckerPose~\cite{lian2023checkerpose} & 35.7 & 84.5 & 97.1 & 79.7 & 98.9 & 79.2 \\
    HccePose~\cite{Wang_2025_ICCV} & 36.8 & 76.6 & 94.4 & 65.7 & 97.9 & 74.3 \\
    VAPO~\cite{lian2024vapo} & \underline{49.2} & \underline{86.1} & 97.0 & -- & -- & -- \\
    Cov2Pose~\cite{Ousalah_2026_CVPR} & -- & -- & \underline{97.2} & -- & -- & -- \\
    Ours & \textbf{56.9} & \textbf{89.5} & \textbf{98.3} & \textbf{86.8} & \textbf{99.4} & \textbf{86.2} \\ \bottomrule
  \end{tabular}
\end{table}

\begin{table}[ht]
  \centering
  \refstepcounter{table}
  \begin{justify}
    \small Table \Roman{table}: Comparison with SOTA methods on the LM-O~\cite{LMO} dataset.
    We highlight the best results in \textbf{bold} and \underline{underline} the second-best results. ``--'' denotes unavailable results.
  \end{justify}
  \label{table:Comparison_lmo}
  \setlength{\tabcolsep}{11.0pt}
  \begin{tabular}{lcccc}
    \toprule
    \multirow{2}{*}{Method} & \multicolumn{3}{c}{ADD(-S)} & \multirow{2}{*}{Mean} \\ \cmidrule(lr){2-4}
    & 0.02d & 0.05d & 0.1d & \\ \midrule
    GDR-Net~\cite{wang2021gdr} & 4.4 & 31.1 & 62.2 & 32.6 \\
    Cov2Pose~\cite{Ousalah_2026_CVPR} & - & - & 76.8 & - \\
    ZebraPose~\cite{su2022zebrapose} & \underline{9.8} & 44.6 & 76.9 & 43.8 \\
    ZebraPose-LC~\cite{Liu_2023_ICCV} & 8.6 & 44.2 & 78.1 & 43.6 \\
    CheckerPose~\cite{lian2023checkerpose} & 7.3 & 43.5 & 77.5 & 42.8 \\
    6D-Diff~\cite{xu20246d} & -- & -- & \textbf{79.6} & -- \\
    VAPO~\cite{lian2024vapo} & 9.7 & 46.2 & 78.0 & 44.6 \\
    HccePose~\cite{Wang_2025_ICCV} & 9.7 & \underline{47.1} & 78.5 & \underline{45.1} \\
    Ours & \textbf{11.2} & \textbf{49.0} & \underline{79.2} & \textbf{46.5} \\ \bottomrule
  \end{tabular}
\end{table}

\subsection{Comparison with State-of-the-Art Methods}
\label{sec:Comparisonsota}
In this section, we compare GHLAPose with state-of-the-art methods on LM~\cite{LM}, LM-O~\cite{LMO}, YCB-V~\cite{YCB-V}, and TUD-L~\cite{hodan2018bop}.
Besides standard pose estimation metrics, we also report BOP results and keypoint localization accuracy to directly evaluate the quality of the predicted correspondences.
Finally, qualitative comparisons with VAPO~\cite{lian2024vapo} further illustrate the improvements in both keypoint localization and pose estimation.

\textbf{Experiments on the LM Dataset.} Table~\ref{table:Comparison_lm} reports the comparison on LM.
Since HccePose~\cite{Wang_2025_ICCV} does not provide results under the LM setting, we retrain a single model for all 13 objects for a fair comparison.
Although ADD(-S) 0.1d on LM is close to saturation, our method still improves the best previous result from 97.2\% to 98.3\%.
More importantly, the gains become much clearer under fine-grained metrics.
Compared with the second-best method, our method improves ADD(-S) 0.02d by 7.7\% and ADD(-S) 0.05d by 3.4\%.
It also achieves 86.8\% on the strict $2^\circ$2cm metric and 86.2\% mean accuracy, outperforming previous methods with available results by 5.8\% on both metrics.
These results demonstrate the effectiveness of the proposed HGB encoding and global-to-local keypoint localization strategy for accurate pose estimation.

\begin{table}[!t]
  \centering
  \refstepcounter{table}
  \begin{justify}
    \small Table \Roman{table}: Comparison with SOTA methods on the YCB-V~\cite{YCB-V} dataset. AUC values computed without and with 11-point interpolation are denoted as ``w/o'' and ``w/ IT'', respectively. ``--'' denotes unavailable results and (*) indicates models trained without the special 80k data mentioned in~\cite{ranftl2025mvtop}.
  \end{justify}
  \label{tab:3}
  \setlength{\tabcolsep}{7.5pt}
  \begin{tabular}{lccccc}
    \toprule
    \multirow{2}{*}{Method} & \multirow{2}{*}{ADD(-S)} & \multicolumn{2}{c}{AUC-ADD-S} & \multicolumn{2}{c}{AUC-ADD(-S)} \\ \cmidrule(l{2pt}r{2pt}){3-4} \cmidrule(l{2pt}r{2pt}){5-6}
    & & w/o & w/ IT & w/o & w/ IT \\ \midrule
    SO-Pose$^{*}$~\cite{di2021so} & 56.8 & -- & 90.9 & -- & 83.9 \\
    GDR-Net$^{*}$~\cite{wang2021gdr} & 60.1 & -- & \underline{91.6} & -- & \underline{84.4} \\
    Cov2Pose$^{*}$~\cite{Ousalah_2026_CVPR} & \underline{69.7} & -- & 90.0 & -- & 82.2 \\
    Ours$^{*}$ & \textbf{80.7} & \textbf{91.8} & \textbf{95.9} & \textbf{87.1} & \textbf{91.8} \\ \midrule
    ZebraPose~\cite{su2022zebrapose} & 80.5 & 90.1 & -- & 85.3 & -- \\
    CheckerPose~\cite{lian2023checkerpose} & 81.4 & 91.3 & 95.3 & 86.4 & 91.1 \\
    ZebraPose-LC~\cite{Liu_2023_ICCV} & 82.4 & 90.8 & 95.0 & 86.1 & 90.8 \\
    6D-Diff~\cite{xu20246d} & 83.8 & 91.5 & -- & 87.0 & -- \\
    VAPO~\cite{lian2024vapo} & \underline{84.9} & \underline{92.3} & \underline{96.4} & \underline{87.9} & \underline{92.7} \\
    Ours & \textbf{85.9} & \textbf{93.1} & \textbf{96.8} & \textbf{88.9} & \textbf{93.9} \\ \bottomrule
  \end{tabular}
\end{table}

\begin{table}[!t]
  \centering
  \refstepcounter{table}
  \begin{justify}
    \small Table \Roman{table}: Results on the YCB-V~\cite{YCB-V} dataset under the BOP setup~\cite{hodavn2020bop}. Results of Coupled~\cite{lipson2022coupled}, NCF~\cite{huang2022neural}, CheckerPose~\cite{lian2023checkerpose}, and HccePose~\cite{Wang_2025_ICCV} are from the original papers, while other results are from the BOP leaderboard. ``--'' denotes unavailable results and (*) denotes inference with 8 views.
  \end{justify}
  \label{tab:ycbv_bop}
  \setlength{\tabcolsep}{9.3pt}
  \begin{tabular}{lcccc}
    \toprule
    Method & MSPD & MSSD & VSD & AR$_{\text{BOP}}$ \\ \midrule
    Coupled~\cite{lipson2022coupled} & 85.2 & 83.5 & 78.3 & 82.4 \\
    ZebraPose~\cite{su2022zebrapose} & 86.4 & 83.0 & 75.1 & 81.5 \\
    ZebraPose-SAT~\cite{su2022zebrapose} & 87.5 & 84.5 & 76.5 & 82.8 \\
    NCF~\cite{huang2022neural} & -- & -- & -- & 77.5 \\
    PFA~\cite{hu2022perspective} & 84.9 & 81.4 & 75.8 & 80.7 \\
    GDRNPP~\cite{gdrnpp} & 86.9 & 84.6 & 76.0 & 82.5 \\
    CheckerPose~\cite{lian2023checkerpose} & 85.3 & 84.4 & 70.7 & 80.1 \\
    CosyPose$^{*}$~\cite{labbe2020cosypose} & 88.0 & \textbf{88.5} & \underline{79.5} & \underline{85.3} \\
    HccePose~\cite{Wang_2025_ICCV} & \underline{88.7} & 85.3 & 77.6 & 83.9 \\
    Ours & \textbf{90.5} & \underline{87.5} & \textbf{79.8} & \textbf{85.9} \\ \bottomrule
  \end{tabular}
\end{table}

\textbf{Experiments on the LM-O Dataset.} Table~\ref{table:Comparison_lmo} reports results on LM-O~\cite{LMO}, where severe occlusion makes reliable correspondence localization more challenging.
Our method achieves the best results on ADD(-S) 0.02d, ADD(-S) 0.05d, and the mean score, improving over the second-best results by 1.4\%, 1.9\%, and 1.4\%, respectively.
The gains are more pronounced compared with closely related keypoint encoding methods.
Compared with CheckerPose~\cite{lian2023checkerpose}, our method improves ADD(-S) 0.02d, 0.05d, and the mean score by 3.9\%, 5.5\%, and 3.7\%.
Compared with VAPO~\cite{lian2024vapo}, it improves the same metrics by 1.5\%, 2.8\%, and 1.9\%, while using less than half of VAPO's 700k training iterations on LM-O.

Recent Cov2Pose~\cite{Ousalah_2026_CVPR} introduces spatial covariance and SPD-manifold modeling into direct pose regression, improving the direct regression paradigm under occlusion.
However, without explicit 2D--3D correspondence localization, it reports 76.8\% on ADD(-S) 0.1d, while our method reaches 79.2\%, a 2.4\% improvement.
Our result is also close to the diffusion-based 6D-Diff~\cite{xu20246d} on ADD(-S) 0.1d (79.2\% $vs.$ 79.6\%), while achieving stronger fine-grained performance and requiring a much lighter inference pipeline.

\begin{figure*}[!t]
  \centering
  \refstepcounter{figure}
  \includegraphics[width=0.99\linewidth]{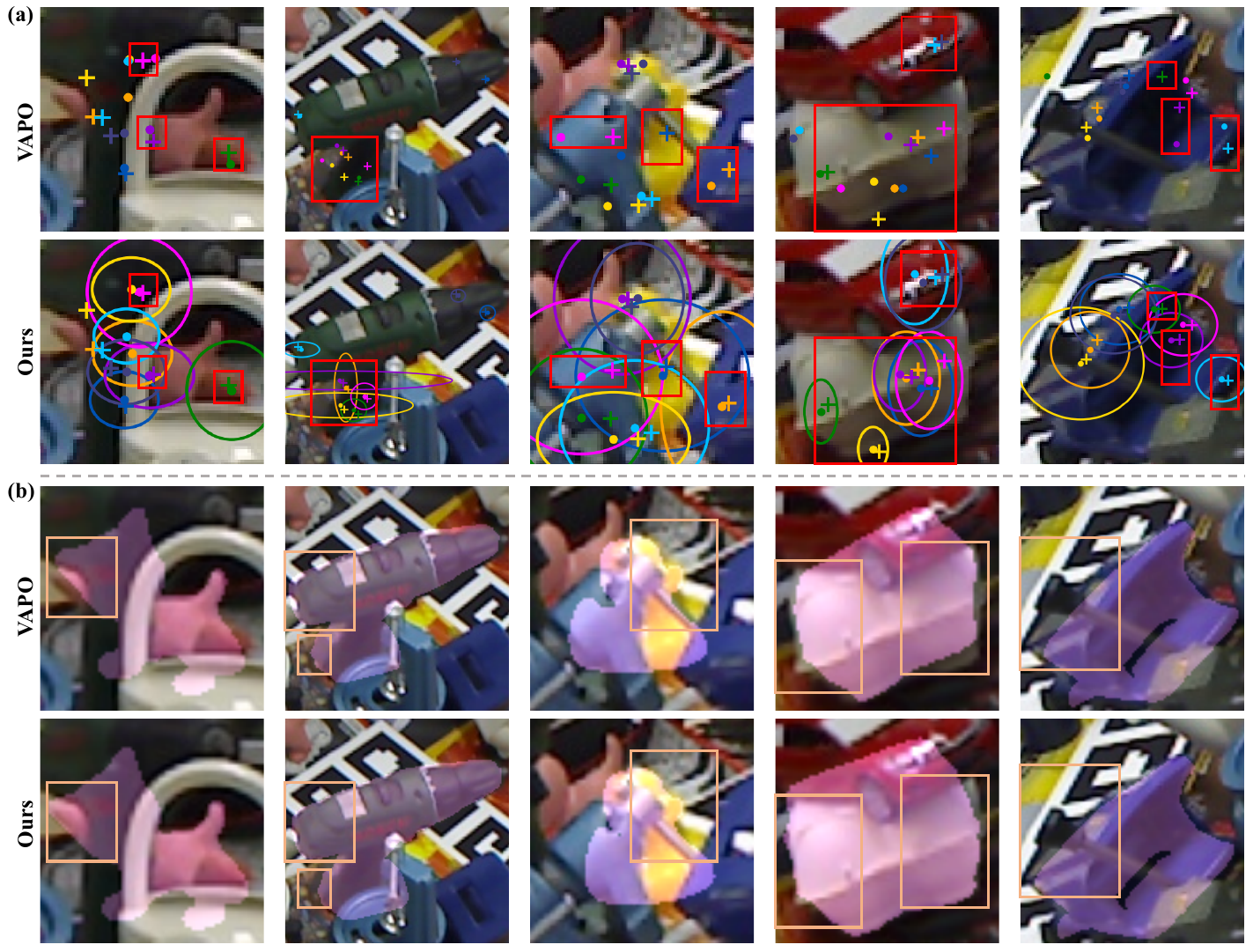}
  \label{fig:comparison_vapo}
  \vspace{-3pt}
  \begin{justify}
    \small \textbf{Fig. \arabic{figure}: Qualitative comparison with VAPO~\cite{lian2024vapo}.}
    \textbf{(a)} Circles denote predicted keypoints, while crosses indicate ground truths.
    The \textcolor{red}{red} arrows highlight the superior keypoint localization achieved by our method.
    \textbf{(b)} We also present the corresponding pose estimation results.
    The \textcolor{myorange}{orange} boxes highlight that our method achieves more accurate pose estimation, consistent with its improvements on fine-grained metrics.
  \end{justify}
\end{figure*}

\textbf{Experiments on the YCB-V Dataset.} Considering the special 80k training data described in~\cite{ranftl2025mvtop}, we conduct separate comparisons with and without this data to ensure a fair evaluation.
Table~\ref{tab:3} reports ADD(-S) 0.1d and the AUC metrics on YCB-V.
Without the special 80k data, our method already achieves 80.7\% ADD(-S), outperforming Cov2Pose~\cite{Ousalah_2026_CVPR} by 11.0\% and GDR-Net~\cite{wang2021gdr} by 20.6\%.
It also improves AUC-ADD-S and AUC-ADD(-S) with interpolation over GDR-Net by 4.3\% and 7.4\%, respectively.
With the full training data, our method further achieves the best results across all metrics.
Compared with the diffusion-based 6D-Diff~\cite{xu20246d}, our method improves ADD(-S) by 2.1\% and AUC-ADD(-S) without interpolation by 1.9\%, while avoiding iterative diffusion refinement.
Compared with VAPO~\cite{lian2024vapo}, our method consistently leads across the reported metrics. 
For example, it improves ADD(-S) by 1.0\% and AUC-ADD(-S) without interpolation by 1.0\%.

Since HccePose~\cite{Wang_2025_ICCV} does not report the standard YCB-V ADD metrics, we further evaluate under the BOP setup~\cite{hodavn2020bop} in Table~\ref{tab:ycbv_bop}.
Our method achieves the best AR$_{\text{BOP}}$ of 85.9\%, surpassing HccePose by 2.0\%.
It also improves MSPD and VSD over HccePose by 1.8\% and 2.2\%, respectively.
Notably, our single-view method also outperforms the multi-view CosyPose~\cite{labbe2020cosypose}, which uses 8 views, by 0.6\% on AR$_{\text{BOP}}$ and 0.3\% on VSD.

\textbf{Experiments on the TUD-L Dataset.} Table~\ref{table:Comparison_tudl} shows that our method achieves the best results across all three ADD(-S) thresholds on TUD-L~\cite{hodan2018bop}.
While HccePose~\cite{Wang_2025_ICCV} constructs dense correspondences from front and back object surfaces, our method refines predefined keypoint locations through a global-to-local strategy.
Compared with HccePose, our method improves ADD(-S) 0.1d by 4.2\% and the mean score by 2.7\%.
These results further demonstrate the effectiveness of our method for pose estimation under complex lighting conditions.

These results demonstrate the applicability of our HGB-based framework across datasets with different object appearances, occlusion levels, and lighting conditions.
HGB represents image-grid coordinates through a common encoding rule that addresses local label discontinuity while restricting decoding dependencies.
LM and LM-O evaluate localization under different degrees of occlusion, YCB-V extends the evaluation to a broader collection of objects, and TUD-L further tests performance under complex lighting conditions.
The results under the respective dataset training settings support the use of this encoding strategy across diverse objects and imaging conditions.
We provide per-object results on LM-O~\cite{LMO}, YCB-V~\cite{YCB-V}, and TUD-L~\cite{hodan2018bop} in Appendices~\ref{sec:supp_lmo_perobj}--\ref{sec:supp_tudl_perobj}.

\begin{table}[ht]
  \centering
  \refstepcounter{table}
  \begin{justify}
    \small Table \Roman{table}: Comparison with SOTA methods on the TUD-L~\cite{hodan2018bop} dataset.
    We highlight the best results in \textbf{bold} and \underline{underline} the second-best results. Mean averages the three ADD(-S) scores.
  \end{justify}
  \label{table:Comparison_tudl}
  \setlength{\tabcolsep}{11.5pt}
  \begin{tabular}{lcccc}
    \toprule
    \multirow{2}{*}{Method} & \multicolumn{3}{c}{ADD(-S)} & \multirow{2}{*}{Mean} \\ \cmidrule(lr){2-4}
    & 0.02d & 0.05d & 0.1d & \\ \midrule
    PFA-CosyPose~\cite{hu2022perspective} & 13.3 & 35.5 & 60.3 & 36.4 \\
    GDRNPP~\cite{gdrnpp} & 12.2 & 36.3 & 63.0 & 37.2 \\
    ZebraPose~\cite{su2022zebrapose} & \underline{19.3} & 44.8 & 64.0 & 42.7 \\
    RACE-6D~\cite{Ha_2026_CVPR} & 12.0 & 39.3 & 69.8 & 40.4 \\
    HccePose~\cite{Wang_2025_ICCV} & 18.0 & \underline{47.3} & \underline{73.3} & \underline{46.2} \\
    Ours & \textbf{20.3} & \textbf{49.0} & \textbf{77.5} & \textbf{48.9} \\ \bottomrule
  \end{tabular}
\end{table}

\vspace{-2mm}
\begin{table*}[!t]
  \centering
  \refstepcounter{table}
  \begin{justify}
    \small Table \Roman{table}: Comparison of keypoint localization on LM~\cite{LM}, LM-O~\cite{LMO}, and YCB-V~\cite{YCB-V} datasets. We reproject the 3D keypoints using ground-truth poses and compute the error between the predicted and reprojected keypoints. For symmetric objects, we use the equivalent rotation closest to the estimated pose. Poses are estimated using the PnP/RANSAC solver~\cite{lepetit2009ep}. Acc denotes accuracy.
  \end{justify}
  \label{tab:keypoint_localization}
  \setlength{\tabcolsep}{6.0pt}
  \begin{tabular}{lccccccccc}
    \toprule
    \multirow{2}{*}{Method} & \multicolumn{3}{c}{LM~\cite{LM}} & \multicolumn{3}{c}{LM-O~\cite{LMO}} & \multicolumn{3}{c}{YCB-V~\cite{YCB-V}} \\ \cmidrule(lr){2-4} \cmidrule(l{2pt}r{2pt}){5-7} \cmidrule(l{2pt}r{2pt}){8-10}
    & Acc@1 & Acc@2 & Acc@5 & Acc@1 & Acc@2 & Acc@5 & Acc@1 & Acc@2 & Acc@5 \\ \midrule
    CheckerPose~\cite{lian2023checkerpose} & 13.9 & 44.8 & 88.4 & 4.3 & 17.2 & 61.6 & 2.0 & 7.6 & 35.0 \\
    VAPO~\cite{lian2024vapo} & 49.5 & 77.2 & 88.5 & 6.2 & 22.6 & 66.1 & -- & -- & -- \\
    Ours (w/o LAR) & \underline{50.6} & \underline{77.9} & \underline{89.4} & \underline{7.8} & \underline{25.3} & \underline{67.0} & \underline{5.7} & \underline{20.0} & \underline{59.5} \\
    Ours & \textbf{58.4} & \textbf{81.5} & \textbf{91.8} & \textbf{8.0} & \textbf{25.9} & \textbf{67.3} & \textbf{6.2} & \textbf{21.2} & \textbf{61.0} \\ \bottomrule
  \end{tabular}
\end{table*}

\textbf{Keypoint Localization Accuracy.} Table~\ref{tab:keypoint_localization} directly evaluates keypoint localization accuracy against CheckerPose~\cite{lian2023checkerpose} and VAPO~\cite{lian2024vapo} on LM, LM-O, and YCB-V.
Although VAPO filters unreliable keypoints using its visibility-aware selection, our method uses all 512 predefined keypoints and still achieves higher localization accuracy.
The improvements are most evident under fine-grained thresholds.
On LM, our method improves Acc@1 over VAPO~\cite{lian2024vapo} by 8.9\%, indicating that HGB provides a more accurate global localization than the binary encoding used in previous dense keypoint methods.
On LM-O, the gain on Acc@2 reaches 3.3\%, showing that this advantage remains under heavy occlusion.
The comparison with the variant without LAR further validates the role of local refinement: LAR improves Acc@1 from 50.6\% to 58.4\% on LM and from 5.7\% to 6.2\% on YCB-V.
These results show that HGB improves the initial global keypoint localization, while LAR further corrects fine-grained local errors.
This validates the proposed global-to-local keypoint localization strategy for establishing more reliable 2D--3D correspondences.

\subsection{Qualitative Results}
Fig.~\ref{fig:comparison_vapo} qualitatively demonstrates the advantage of our method from both keypoint localization and pose estimation perspectives.
In Fig.~\ref{fig:comparison_vapo}a, VAPO~\cite{lian2024vapo} often exhibits visible keypoint deviations from the ground truth, especially under occlusion and clutter.
These errors are mainly caused by discrete biases from binary encoding and the remaining quantization errors after decoding.
In contrast, HGB provides more reliable global localization, and LAR further refines the coarse estimates into accurate local coordinates.
The improved keypoint localization directly benefits pose estimation.
As shown in Fig.~\ref{fig:comparison_vapo}b, the localization errors of VAPO are amplified through pose solving and lead to visible pose deviations, while our method produces more accurate alignments in challenging occluded scenes.

\subsection{Runtime Analysis}
\label{sec:runtime}
Table~\ref{tab:5} reports network complexity and runtime on the LM-O dataset.
GFLOPs and parameter counts cover the correspondence prediction network.
Following~\cite{lian2024vapo}, runtime includes correspondence prediction and pose solving, excluding object detection.
All methods are evaluated on an Intel Ultra 9 285K CPU and an NVIDIA RTX 5090 GPU.
Our network requires 153.2 GFLOPs and 25.1 M parameters to establish high-quality correspondences from 512 predefined keypoints.
HGB improves global keypoint localization, and LAR corrects residual errors without increasing the correspondence count, leaving at most 512 correspondences for pose initialization after filtering.
With this compact correspondence set, PnP/RANSAC~\cite{lepetit2009ep} initialization and uncertainty-weighted Ceres~\cite{Agarwal_Ceres_Solver_2022} refinement together take 1.0 ms.
Compared with VAPO~\cite{lian2024vapo}, which uses Progressive-X~\cite{barath2019progressive}, our network requires more computation but adds only 0.7 ms to correspondence prediction (15.7 ms vs.\ 15.0 ms), while reducing pose-solving time from 4.3 ms to 1.0 ms.
The saving in pose solving offsets the additional network inference time, yielding the lowest total runtime among the compared methods at 16.7 ms.
This low latency, together with accurate pose estimation, highlights the potential of our method for real-world robotic applications.
\begin{table}[ht]
  \centering
  \refstepcounter{table}
  \begin{justify}
    \small Table \Roman{table}: Network complexity and runtime comparison. All runtimes are in ms. (*) denotes methods using Progressive-X~\cite{barath2019progressive}. Otherwise PnP/RANSAC~\cite{lepetit2009ep} is used. Solver settings correspond to each method's best performance in its original paper.
  \end{justify}
  \label{tab:5}
  \setlength{\tabcolsep}{1.6pt}
  \begin{tabular}{lccccc}
    \toprule
    \multirow[c]{2}{*}[-1pt]{Method} & \multicolumn{3}{c}{Network (Correspondence prediction)} & \multirow[c]{2}{*}[-1pt]{Pose solve} & \multirow[c]{2}{*}[-1pt]{Total} \\
    \cmidrule(lr){2-4}
     & GFLOPs & Params. (M) & Inference & & \\
    \midrule
    ZebraPose$^{*}$~\cite{su2022zebrapose} & 90.9 & 41.5 & \textbf{2.5} & 86.8 & 89.3 \\
    CheckerPose$^{*}$~\cite{lian2023checkerpose} & \textbf{55.4} & \textbf{21.3} & 13.4 & 6.0 & 19.4 \\
    VAPO$^{*}$~\cite{lian2024vapo} & 98.7 & 23.6 & 15.0 & 4.3 & 19.3 \\
    HccePose~\cite{Wang_2025_ICCV} & 109.5 & 29.1 & 6.8 & 29.9 & 36.7 \\
    Ours & 153.2 & 25.1 & 15.7 & \textbf{1.0} & \textbf{16.7} \\
    \bottomrule
  \end{tabular}
\end{table}

\section{Conclusion}
\label{sec:Conclusion}
In this paper, we propose GHLAPose, which consists of an HGB keypoint representation and LAR.
The HGB representation mitigates \textit{large transitions} in simple binary, reducing training difficulty and thereby enhancing global keypoint localization.
LAR further alleviates the discrete and quantization errors from encoding, forming a global-to-local keypoint localization framework while estimating uncertainty.
Subsequently, we estimate the object pose using a weighted PnP solver.
Experiments on the LM, LM-O, YCB-V, and TUD-L datasets demonstrate that our GHLAPose achieves SOTA pose estimation performance while also attaining the fastest inference speed.

\textbf{Limitations and Future Work.} GHLAPose achieves strong performance through a two-stage global-to-local keypoint localization framework.
However, its training currently involves two separate stages, and the performance of LAR depends on the preceding CodeNet.
Future work may develop a unified training strategy and a corresponding network architecture to jointly optimize encoding-based global localization and regression-based local refinement.
Extending the single-image framework with temporal consistency and uncertainty-guided tracking could improve stability in dynamic scenes with changing visibility.
Another promising direction is to extend the proposed global-to-local localization strategy to more general category-level~\cite{Diff-COPE,11251041,CLIPose} pose estimation, where object instances and shapes are not fixed during training.

\renewcommand*{\bibfont}{\footnotesize}
\printbibliography

\begin{IEEEbiography}[{\includegraphics[width=1in,height=1.25in,clip,keepaspectratio]{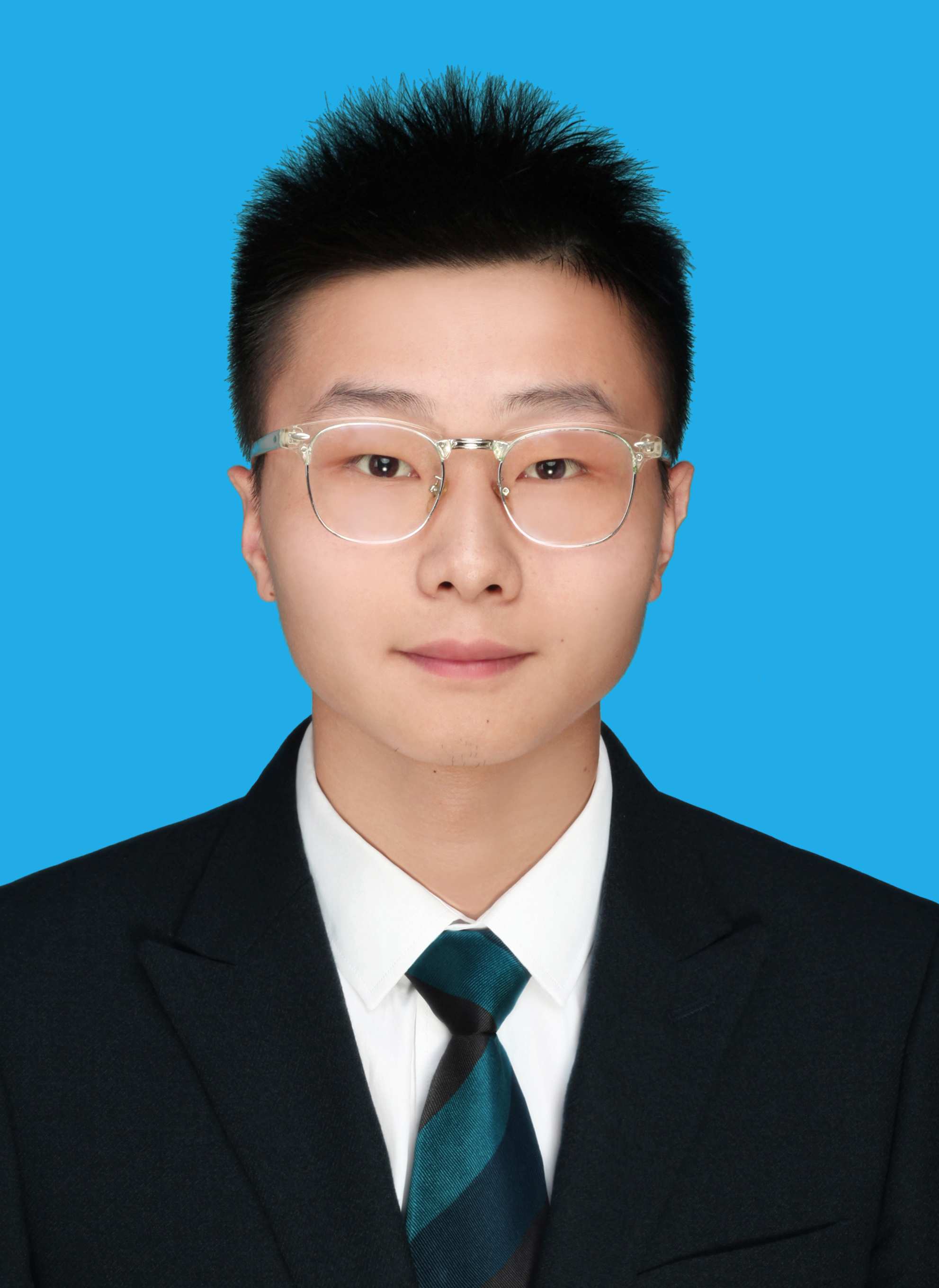}}]{Jiahong Chen}
  received the B.E. degree in Mechanical Design, Manufacturing and Automation from Wuhan University of Technology in 2023, and the M.E. degree in Mechanical Engineering from Wuhan University in 2025. He is currently pursuing the Ph.D. degree in Aeronautical and Astronautical Science and Technology at the National University of Defense Technology. His research interests include 3D computer vision and object pose estimation.
\end{IEEEbiography}

\begin{IEEEbiography}[{\includegraphics[width=1in,height=1.25in,clip,keepaspectratio]{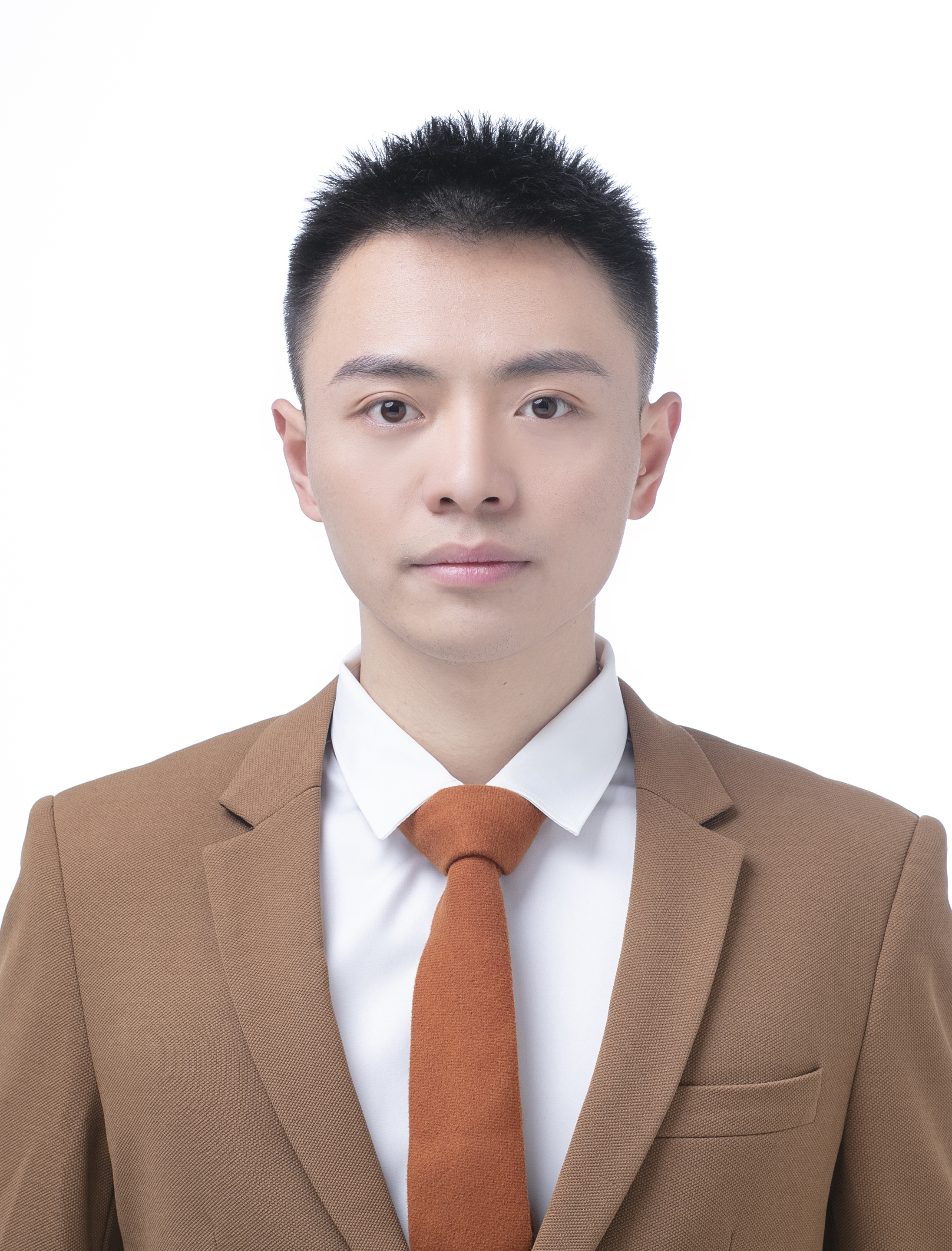}}]{Jinghao Wang}
  received B.E. and ME. degrees from National University of Defense Technology (NUDT), Changsha, China, in 2020 and 2022, respectively. He is currently pursuing a Ph.D. in Aeronautical and Astronautical Science and Technology from National University of Defense Technology (NUDT), Changsha, China. His research interests include satellite pose estimation and object detection in remote sensing scene.
\end{IEEEbiography}

\begin{IEEEbiography}
	[{\includegraphics[width=1in,height=1.25in,clip,keepaspectratio]{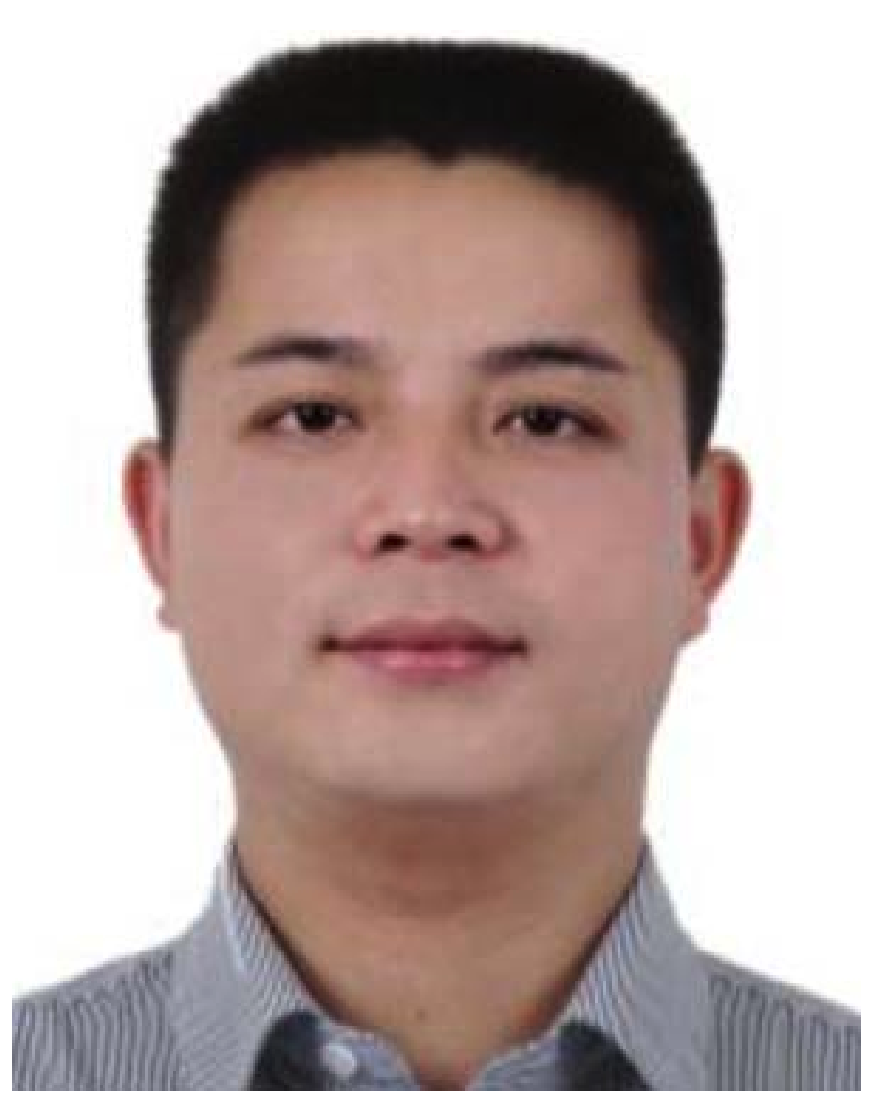}}]{Yueneng Yang} 
	was born in Shaoyang City, Hunan Province, China, in 1984. He received the B.S. and M.S. degrees in civil engineering and the Ph.D. degree in aeronautical and astronautical science and technology from the National University of Defense Technology, Changsha, China, in 2006, 2008, and 2013, respectively.

  He is currently an Associate Professor with the College of Aerospace Science and Engineering, National University of Defense Technology. He authored or coauthored more than 40 papers in journals and conference proceedings. His research interests include bionic flying robots and flight control.
\end{IEEEbiography}

\begin{IEEEbiography}
	[{\includegraphics[width=1in,height=1.25in,clip,keepaspectratio]{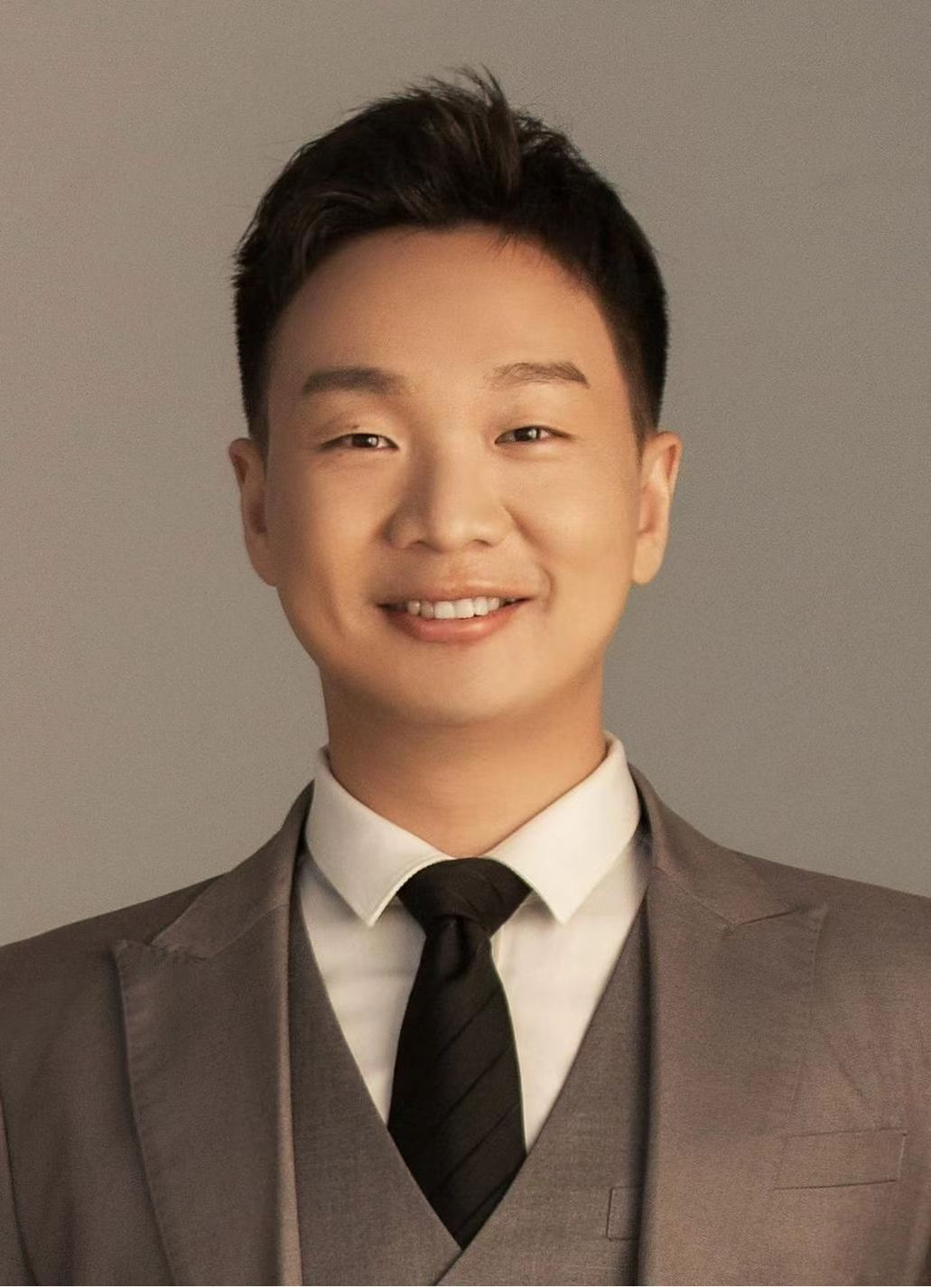}}]{Banglei Guan} 
	received the B.S. degree in geomatics engineering from Wuhan University, Wuhan, China, in 2012 and the Ph.D. degree in aeronautical and astronautical science and technology from the National University of Defense Technology, Changsha, China, in 2018. From 2016 to 2017, he was an Exchange Student with the Graz University of Technology, Graz, Austria.
	He is currently an Associate Professor with the National University of Defense Technology. His research interests include computer vision and photogrammetry.
\end{IEEEbiography}

\begin{IEEEbiography}[{\includegraphics[width=1in,height=1.25in,clip,keepaspectratio]{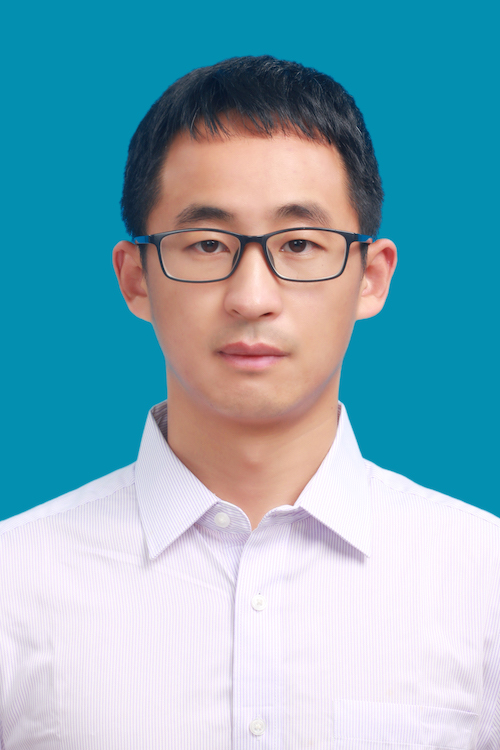}}]{Zi Wang}
	received a B.E. degree in measuring and controlling technologies and instruments from Tianjin University (TJU), Tianjin, China, in 2016,
	and M.E. and Ph.D. degrees in Aeronautical and Astronautical Science and Technology from National University of Defense Technology (NUDT),
        Changsha, China, in 2018 and 2022, respectively.
	He was a visiting Ph.D. student in Sun Yat-sen University from 2021 to 2022.

	His research interests include the development of machine learning techniques and navigation algorithms for spaceborne computer vision tasks,
specifically on robust and accurate estimation of the relative position and attitude of space objects and UAV using monocular vision and deep learning.
He won the 1st place on the sunlamp task of the second Satellite Pose Estimation Competition (SPEC2021).
\end{IEEEbiography}

\begin{IEEEbiography}[{\includegraphics[width=1in,height=1.25in,clip,keepaspectratio]{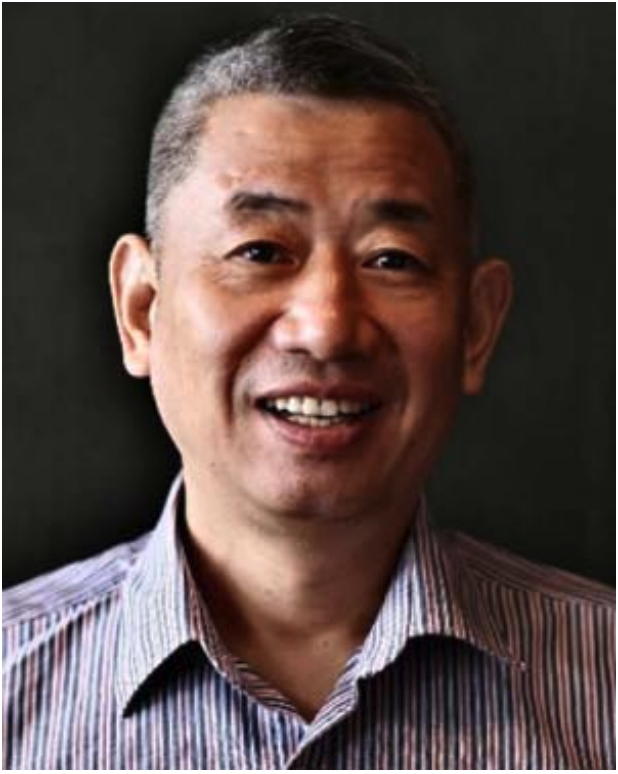}}]{Dewen Hu}
 (Senior Member, IEEE) received the BS and MS degrees from Xi’an Jiaotong University, Xi’an, China, in 1983 and 1986, respectively, and the PhD degree from the National University of Defense Technology, Changsha, China, in 1999. He has visited The University of Sheffield, Sheffield, U.K., from 1995 to 1996. He is currently a professor with the National University of Defense Technology. He has more than 400 publications in journals and conferences such as Proceedings of the National Academy of Sciences of the United States of America, IEEE Transactions on Pattern Analysis and Machine Intelligence, and International Journal of Computer Vision, focusing on pattern recognition and cognitive science. Dr. Hu serves as an associate editor for IEEE T-SMCS.
\end{IEEEbiography}

\begin{IEEEbiography}[{\includegraphics[width=1in,height=1.25in,clip,keepaspectratio]{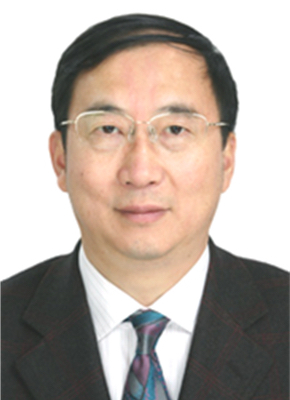}}]{Qifeng Yu}
    received a B.S. degree from Northwestern Polytechnic University, Xi'an, China, in 1981,
    a M.S. degree from the National University of Defense Technology, Changsha, China, in 1984,
    and a Ph.D. from Bremen University, Bremen, Germany, in 1996.
    He is a Professor with the National University of Defense Technology.
    He has authored three books and published over 100 articles.
    His main research fields include image measurement, vision navigation, and close-range photogrammetry.
    Dr. Yu is a member of the Chinese Academy of Sciences.
\end{IEEEbiography}

\clearpage
\twocolumn[{\centering\Large\bfseries Supplementary Material\par\vspace{1em}}]
\appendices
\setlength{\dblfloatsep}{9pt}
\setcounter{figure}{0}
\setcounter{table}{0}
\renewcommand{\thefigure}{S\arabic{figure}}
\renewcommand{\thetable}{S\arabic{table}}
\renewcommand{\theHfigure}{supp.\arabic{figure}}
\renewcommand{\theHtable}{supp.\arabic{table}}
\section{Hyper-parameters in the Pose Solver}
\label{sec:supp_solver}
On the LM~\cite{LM} dataset, we use Progressive-X~\cite{barath2019progressive}, set the reprojection error threshold to 2 pixels, and perform 400 iterations.
The additional parameters for Progressive-X are set as neighborhood ball radius $=30$ and spatial coherence weight $=0.1$.
For PnP/RANSAC~\cite{lepetit2009ep}, we set the reprojection error threshold to 2 pixels on LM and YCB-V~\cite{YCB-V}, and 1.5 pixels on LM-O~\cite{LMO}, with 150 iterations.

\section{Filtering with Mask on LM-O and YCB-V}
\label{sec:supp_filtering}
We use the visible mask $M_{vis}$ to filter unreliable correspondences for selected objects.
On LM-O~\cite{LMO}, $M_{vis}$ is applied to ape, can, cat, driller, eggbox, and duck.
On YCB-V~\cite{YCB-V}, it is applied to master\_chef\_can, potted\_meat\_can, pudding\_box, and foam\_brick.

\section{Additional Ablation Study Results}
\subsection{Additional Results on HGB Encoding}
\label{sec:supp_hgb}
\textbf{Analysis of Encoding Errors.}
Table~\ref{tab:app_bit_error} reports the bit-wise prediction errors of simple binary code and HGB code on the LM~\cite{LM} dataset.
Because HGB keeps the high-order bits unchanged and applies Gray coding only to the last two bits, the errors of the first five bits remain nearly identical to simple binary coding.
In contrast, HGB clearly reduces the errors of the low-order bits, especially the final bit, where the error decreases from 39.4 to 23.7 on the $x$ axis and from 37.4 to 21.0 on the $y$ axis.
This supports our analysis that HGB reduces local encoding discontinuity and eases fine-grained code prediction.

\begin{table*}[!t]
  \centering
  \refstepcounter{table}
  \begin{justify}
    \small Table \thetable: Bit-error comparison between simple binary and HGB code on the LM~\cite{LM} dataset. The values are reported in units of $10^{-2}$. We highlight the best results in \textbf{bold}.
  \end{justify}
  \label{tab:app_bit_error}
  \setlength{\tabcolsep}{3.4pt}
  \begin{tabular}{lcccccccccccccc}
    \toprule
    \multirow{2}{*}{Encoding Method} & \multicolumn{2}{c}{Bit 1} & \multicolumn{2}{c}{Bit 2} & \multicolumn{2}{c}{Bit 3} & \multicolumn{2}{c}{Bit 4} & \multicolumn{2}{c}{Bit 5} & \multicolumn{2}{c}{Bit 6} & \multicolumn{2}{c}{Bit 7} \\ \cmidrule(lr){2-3} \cmidrule(lr){4-5} \cmidrule(lr){6-7} \cmidrule(lr){8-9} \cmidrule(lr){10-11} \cmidrule(lr){12-13} \cmidrule(lr){14-15}
    & $x$ & $y$ & $x$ & $y$ & $x$ & $y$ & $x$ & $y$ & $x$ & $y$ & $x$ & $y$ & $x$ & $y$ \\ \midrule
    Simple Binary & \textbf{1.2} & \textbf{1.0} & \textbf{2.1} & \textbf{2.1} & \textbf{4.2} & \textbf{4.1} & \textbf{7.5} & \textbf{7.1} & \textbf{13.6} & \textbf{12.6} & 24.5 & 22.2 & 39.4 & 37.4 \\
    HGB Code (Ours) & \textbf{1.2} & \textbf{1.0} & \textbf{2.1} & 2.2 & \textbf{4.2} & 4.2 & \textbf{7.5} & 7.2 & \textbf{13.6} & \textbf{12.6} & \textbf{24.2} & \textbf{21.5} & \textbf{23.7} & \textbf{21.0} \\ \bottomrule
  \end{tabular}
\end{table*}

\begin{figure}[!t]
  \centering
  \refstepcounter{figure}
  \includegraphics[width=\linewidth]{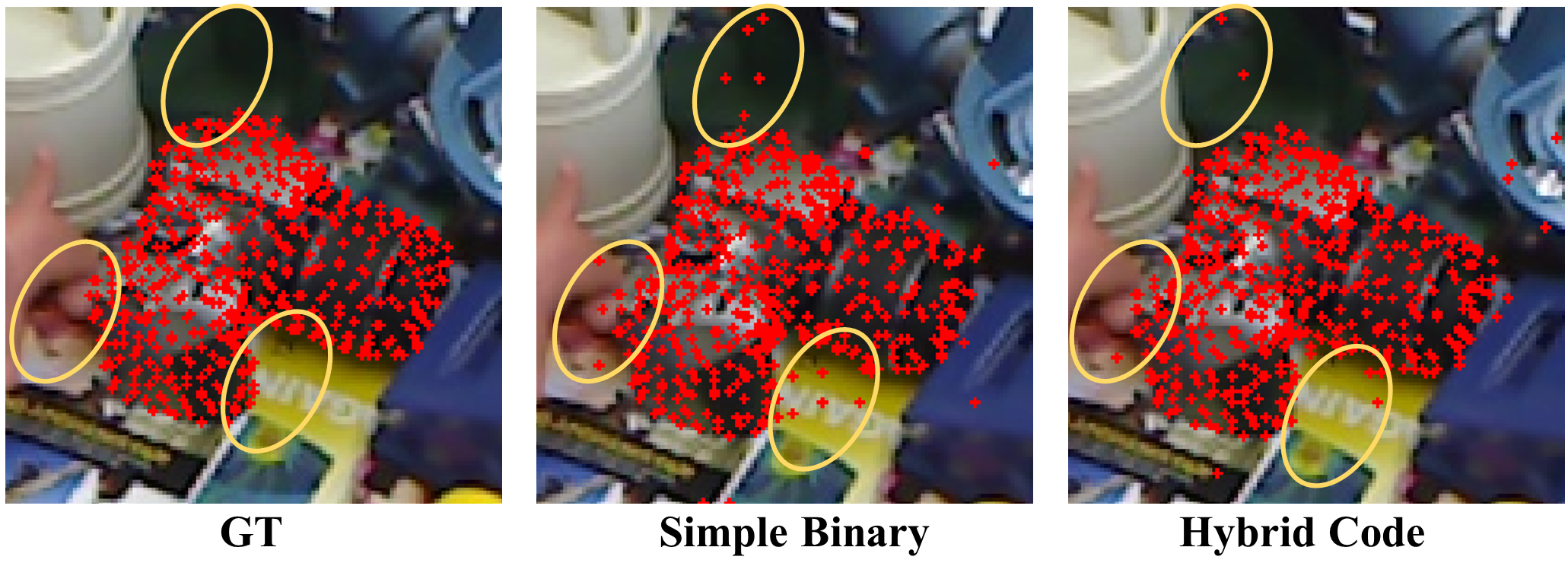}
  \label{fig:supp_camera}
  \vspace{-8pt}
  \begin{justify}
    \small \textbf{Fig. \thefigure: Simple binary and HGB keypoint visualization.}
    We show predicted keypoints for the \textit{camera} object.
    The highlighted regions show that HGB produces fewer outliers than simple binary encoding.
  \end{justify}
\end{figure}

\begin{figure}[!t]
  \centering
  \refstepcounter{figure}
  \includegraphics[width=\linewidth]{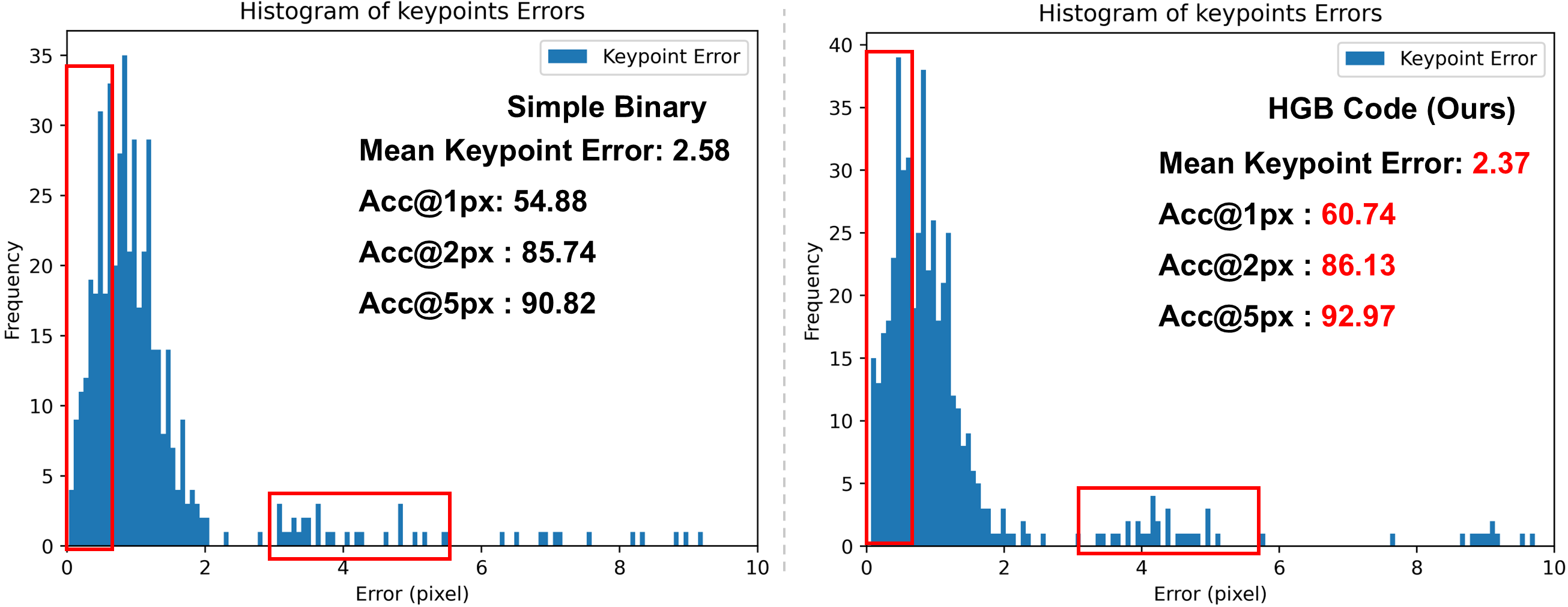}
  \label{fig:supp_kpt_err}
  \vspace{-8pt}
  \begin{justify}
    \small \textbf{Fig. \thefigure: Keypoint error distribution on the \textit{camera} object.}
    HGB reduces keypoint localization errors, especially in the fine-grained range.
  \end{justify}
\end{figure}

\textbf{Qualitative Results and Keypoint Error Distribution.}
Fig.~\ref{fig:supp_camera} visualizes all keypoints predicted by simple binary code and HGB code on the \textit{camera} object.
HGB effectively reduces the number of outliers in global keypoint localization.
Fig.~\ref{fig:supp_kpt_err} further shows the corresponding keypoint error distribution.
The distribution shifts toward smaller errors, especially in the fine-grained range, validating the effectiveness of HGB encoding.

\section{LM-O Evaluation per Object}
\label{sec:supp_lmo_perobj}
Tables~\ref{tab:app_lmo_add002}, \ref{tab:app_lmo_add005}, and \ref{tab:app_lmo_add01} provide object-level results on LM-O~\cite{LMO} under ADD(-S) 0.02d, 0.05d, and 0.1d, respectively.
The fine-grained metrics in Tables~\ref{tab:app_lmo_add002} and \ref{tab:app_lmo_add005} further verify the advantage of the proposed global-to-local keypoint localization strategy.
Under the looser ADD(-S) 0.1d metric, our method remains competitive with the best-performing methods while achieving stronger fine-grained accuracy.

\begin{table*}[!t]
  \centering
  \refstepcounter{table}
  \begin{justify}
    \small Table \thetable: Object-level results on LM-O~\cite{LMO} w.r.t. ADD(-S) 0.02d. We highlight the best results in \textbf{bold} and \underline{underline} the second-best results. ``*'' denotes symmetric objects.
  \end{justify}
  \label{tab:app_lmo_add002}
  \setlength{\tabcolsep}{4.5pt}
  \begin{tabular}{lccccccc}
    \toprule
    Method & GDR-Net~\cite{wang2021gdr} & ZebraPose~\cite{su2022zebrapose} & ZebraPose-LC~\cite{Liu_2023_ICCV} & CheckerPose~\cite{lian2023checkerpose} & VAPO~\cite{lian2024vapo} & HccePose~\cite{Wang_2025_ICCV} & Ours \\ \midrule
    ape & \underline{0.4} & 0.2 & \textbf{0.5} & \textbf{0.5} & \textbf{0.5} & \underline{0.4} & \underline{0.4} \\
    can & 10.8 & 20.7 & 20.9 & 15.6 & \textbf{24.0} & 17.2 & \underline{22.0} \\
    cat & 0.8 & 0.7 & 2.5 & 0.3 & 1.5 & \textbf{3.4} & \underline{2.9} \\
    driller & 12.9 & 19.7 & 20.3 & 10.1 & 20.7 & 21.7 & \textbf{26.2} \\
    duck & 0.5 & 0.2 & 0.1 & 0.5 & 0.4 & \textbf{1.0} & \underline{0.8} \\
    eggbox* & 0.8 & \textbf{9.3} & 2.1 & 5.4 & 3.6 & \underline{8.4} & 7.8 \\
    glue* & 8.6 & \underline{27.3} & 22.1 & 25.7 & 26.6 & 25.4 & \textbf{29.5} \\
    holep. & 0.2 & \textbf{0.6} & 0.2 & 0.2 & 0.2 & \underline{0.4} & 0.3 \\ \midrule
    Mean & 4.4 & 9.8 & 8.6 & 7.3 & 9.7 & 9.7 & \textbf{11.2} \\ \bottomrule
  \end{tabular}
\end{table*}

\begin{table*}[!t]
  \centering
  \refstepcounter{table}
  \begin{justify}
    \small Table \thetable: Object-level results on LM-O~\cite{LMO} w.r.t. ADD(-S) 0.05d. We highlight the best results in \textbf{bold} and \underline{underline} the second-best results. ``*'' denotes symmetric objects.
  \end{justify}
  \label{tab:app_lmo_add005}
  \setlength{\tabcolsep}{4.5pt}
  \begin{tabular}{lccccccc}
    \toprule
    Method & GDR-Net~\cite{wang2021gdr} & ZebraPose~\cite{su2022zebrapose} & ZebraPose-LC~\cite{Liu_2023_ICCV} & CheckerPose~\cite{lian2023checkerpose} & VAPO~\cite{lian2024vapo} & HccePose~\cite{Wang_2025_ICCV} & Ours \\ \midrule
    ape & 14.2 & 14.4 & 16.5 & 14.0 & 16.5 & \textbf{20.0} & \underline{19.5} \\
    can & 66.5 & 79.3 & 81.9 & 79.2 & \underline{82.0} & 75.6 & \textbf{82.4} \\
    cat & 15.7 & 26.0 & 26.3 & 25.1 & 26.6 & \underline{28.7} & \textbf{31.9} \\
    driller & 55.6 & 71.2 & 74.3 & 71.4 & \textbf{78.6} & 72.0 & \underline{76.0} \\
    duck & 13.8 & 15.9 & 16.1 & 19.8 & \underline{20.3} & 19.5 & \textbf{21.8} \\
    eggbox* & 23.2 & 46.2 & 36.8 & 41.4 & 37.5 & \textbf{52.4} & \underline{48.3} \\
    glue* & 40.8 & \textbf{70.5} & 69.0 & 67.1 & 68.3 & \underline{68.8} & 67.2 \\
    holep. & 19.1 & 33.6 & 32.6 & 29.8 & \underline{40.1} & 40.0 & \textbf{45.0} \\ \midrule
    Mean & 31.1 & 44.6 & 44.2 & 43.5 & 46.2 & \underline{47.1} & \textbf{49.0} \\ \bottomrule
  \end{tabular}
\end{table*}

\begin{table*}[!t]
  \centering
  \refstepcounter{table}
  \begin{justify}
    \small Table \thetable: Object-level results on LM-O~\cite{LMO} w.r.t. ADD(-S) 0.1d. We highlight the best results in \textbf{bold} and \underline{underline} the second-best results. ``*'' denotes symmetric objects.
  \end{justify}
  \label{tab:app_lmo_add01}
  \setlength{\tabcolsep}{6.5pt}
  \begin{tabular}{lcccccccc}
    \toprule
    Method & GDR-Net~\cite{wang2021gdr} & ZebraPose~\cite{su2022zebrapose} & ZebraPose-LC~\cite{Liu_2023_ICCV} & CheckerPose~\cite{lian2023checkerpose} & 6D-Diff~\cite{xu20246d} & VAPO~\cite{lian2024vapo} & HccePose~\cite{Wang_2025_ICCV} & Ours \\ \midrule
    ape & 46.8 & 57.9 & \textbf{61.6} & 58.3 & \underline{60.6} & \underline{60.6} & 59.4 & \textbf{61.6} \\
    can & 90.8 & 95.0 & \underline{97.4} & 95.7 & \textbf{97.9} & 96.8 & 96.1 & 96.8 \\
    cat & 40.5 & 60.6 & \underline{64.5} & 62.3 & 63.2 & 63.1 & 61.7 & \textbf{64.8} \\
    driller & 82.6 & 94.8 & 94.7 & 93.7 & \textbf{96.6} & \underline{95.2} & 95.1 & 94.7 \\
    duck & 46.9 & 64.5 & 66.8 & \textbf{69.9} & 67.2 & 66.0 & 67.1 & \underline{68.8} \\
    eggbox* & 54.2 & 70.9 & 71.8 & 70.0 & \underline{73.5} & 70.5 & \textbf{76.7} & 72.7 \\
    glue* & 75.8 & 88.7 & 86.4 & 86.4 & \textbf{92.0} & 86.2 & 88.5 & 86.1 \\
    holep. & 60.1 & 83.0 & 81.5 & 83.8 & 85.5 & 85.7 & 83.0 & \textbf{87.8} \\ \midrule
    Mean & 62.2 & 76.9 & 78.1 & 77.5 & \textbf{79.6} & 78.0 & 78.5 & \underline{79.2} \\ \bottomrule
  \end{tabular}
\end{table*}

\section{YCB-V Evaluation per Object}
\label{sec:supp_ycbv_perobj}
Tables~\ref{tab:app_ycbv_add01} and \ref{tab:app_ycbv_auc} report object-level results on YCB-V~\cite{YCB-V}.
As shown in Table~\ref{tab:app_ycbv_add01}, GHLAPose achieves the best mean ADD(-S) 0.1d score across all objects.
Table~\ref{tab:app_ycbv_auc} further shows that GHLAPose obtains the best mean AUC of ADD-S and ADD(-S), demonstrating consistent improvements across different thresholds and object categories.

\begin{table*}[!t]
  \centering
  \refstepcounter{table}
  \begin{justify}
    \small Table \thetable: Object-level results on YCB-V~\cite{YCB-V} w.r.t. ADD(-S) 0.1d. Since SO-Pose~\cite{di2021so} does not report object-level results, it is omitted. We highlight the best results in \textbf{bold}. ``*'' denotes symmetric objects.
  \end{justify}
  \label{tab:app_ycbv_add01}
  \setlength{\tabcolsep}{4.5pt}
  \begin{tabular}{lccccccc}
    \toprule
    Method & GDR-Net~\cite{wang2021gdr} & ZebraPose~\cite{su2022zebrapose} & CheckerPose~\cite{lian2023checkerpose} & ZebraPose-LC~\cite{Liu_2023_ICCV} & 6D-Diff~\cite{xu20246d} & VAPO~\cite{lian2024vapo} & Ours \\ \midrule
    002\_master\_chef\_can & 41.5 & 62.6 & 45.9 & 51.6 & \textbf{65.9} & 45.8 & 62.2 \\
    003\_cracker\_box & 83.2 & 98.5 & 94.2 & 99.7 & 99.0 & \textbf{99.9} & 99.3 \\
    004\_sugar\_box & 91.5 & 96.3 & 98.3 & 99.4 & 98.6 & \textbf{99.9} & 99.3 \\
    005\_tomato\_soup\_can & 65.9 & 80.5 & 83.2 & 79.6 & 85.9 & 85.3 & \textbf{89.2} \\
    006\_mustard\_bottle & 90.2 & \textbf{100.0} & 99.2 & 99.7 & \textbf{100.0} & \textbf{100.0} & 99.7 \\
    007\_tuna\_fish\_can & 44.2 & 70.5 & 88.9 & 86.1 & 75.4 & 82.3 & \textbf{89.3} \\
    008\_pudding\_box & 2.8 & 99.5 & 86.5 & 99.1 & \textbf{99.3} & 77.1 & 80.4 \\
    009\_gelatin\_box & 61.7 & 97.2 & 86.0 & 94.9 & 98.1 & \textbf{98.6} & 83.6 \\
    010\_potted\_meat\_can & 64.9 & 76.9 & 70.0 & 73.9 & \textbf{80.2} & 72.7 & 76.8 \\
    011\_banana & 64.1 & 71.2 & 96.0 & 95.8 & 83.5 & 93.7 & \textbf{98.9} \\
    019\_pitcher\_base & 99.0 & \textbf{100.0} & \textbf{100.0} & \textbf{100.0} & \textbf{100.0} & \textbf{100.0} & \textbf{100.0} \\
    021\_bleach\_cleanser & 73.8 & 75.9 & \textbf{89.8} & 85.6 & 82.3 & 89.3 & 89.1 \\
    024\_bowl* & 37.7 & 18.5 & 68.0 & 35.2 & 23.2 & \textbf{99.8} & 98.0 \\
    025\_mug & 61.5 & 77.5 & 89.0 & 88.7 & 84.4 & \textbf{95.0} & 73.0 \\
    035\_power\_drill & 78.5 & 97.4 & 95.9 & \textbf{99.2} & 98.4 & 98.6 & 94.7 \\
    036\_wood\_block* & 59.5 & 87.6 & 58.7 & 82.6 & \textbf{89.8} & 63.2 & 81.8 \\
    037\_scissors & 3.9 & 71.8 & 62.4 & 56.9 & \textbf{79.6} & 76.8 & 76.8 \\
    040\_large\_marker & 7.4 & 23.3 & 18.8 & 27.8 & \textbf{28.4} & 25.1 & 20.1 \\
    051\_large\_clamp* & 69.8 & 87.6 & \textbf{95.4} & 84.4 & 90.3 & 89.3 & 93.8 \\
    052\_extra\_large\_clamp* & 90.0 & 98.0 & 95.6 & 99.1 & 98.6 & \textbf{99.3} & 99.0 \\
    061\_foam\_brick* & 71.9 & 99.3 & 87.2 & 91.3 & \textbf{99.5} & 91.3 & 99.3 \\ \midrule
    Mean & 60.1 & 80.5 & 81.4 & 82.4 & 83.8 & 84.9 & \textbf{85.9} \\ \bottomrule
  \end{tabular}
\end{table*}

\begin{table*}[!t]
  \centering
  \refstepcounter{table}
  \begin{justify}
    \small Table \thetable: Object-level results on YCB-V~\cite{YCB-V} w.r.t. AUC of ADD-S and ADD(-S). Symmetric metric is used for all objects in ADD-S and only for symmetric objects in ADD(-S). VAPO~\cite{lian2024vapo} does not report object-level AUC results and is omitted. We highlight the best results in \textbf{bold}. ``*'' denotes symmetric objects.
  \end{justify}
  \label{tab:app_ycbv_auc}
  \setlength{\tabcolsep}{4.5pt}
  \begin{tabular}{lcccccccccc}
    \toprule
    Method & \multicolumn{2}{c}{ZebraPose~\cite{su2022zebrapose}} & \multicolumn{2}{c}{CheckerPose~\cite{lian2023checkerpose}} & \multicolumn{2}{c}{ZebraPose-LC~\cite{Liu_2023_ICCV}} & \multicolumn{2}{c}{6D-Diff~\cite{xu20246d}} & \multicolumn{2}{c}{Ours} \\ \cmidrule(lr){2-3} \cmidrule(lr){4-5} \cmidrule(lr){6-7} \cmidrule(lr){8-9} \cmidrule(lr){10-11}
    Metric & ADD-S & ADD(-S) & ADD-S & ADD(-S) & ADD-S & ADD(-S) & ADD-S & ADD(-S) & ADD-S & ADD(-S) \\ \midrule
    002\_master\_chef\_can & 93.7 & 75.4 & 87.5 & 67.7 & 88.4 & 66.9 & \textbf{94.3} & 77.3 & 91.1 & \textbf{81.0} \\
    003\_cracker\_box & 93.0 & 87.8 & 93.2 & 86.7 & 93.7 & 88.3 & 93.7 & 88.1 & \textbf{94.5} & \textbf{89.7} \\
    004\_sugar\_box & 95.1 & 90.9 & 95.9 & 91.7 & 94.7 & 90.3 & \textbf{96.3} & 91.8 & 95.9 & \textbf{92.5} \\
    005\_tomato\_soup\_can & 94.4 & 90.1 & 94.0 & 89.9 & 93.4 & 89.2 & \textbf{95.4} & 91.3 & \textbf{95.4} & \textbf{92.1} \\
    006\_mustard\_bottle & 96.0 & 92.6 & 95.7 & 90.9 & 95.1 & 90.9 & \textbf{96.5} & \textbf{92.9} & 96.1 & 92.5 \\
    007\_tuna\_fish\_can & 96.9 & 92.6 & 97.5 & 94.4 & 97.2 & 94.1 & 96.9 & 93.8 & \textbf{97.6} & \textbf{94.5} \\
    008\_pudding\_box & 97.2 & 95.3 & 94.9 & 91.5 & 96.7 & 94.7 & \textbf{97.6} & \textbf{95.6} & 94.1 & 89.9 \\
    009\_gelatin\_box & 96.8 & 94.8 & 96.1 & 93.4 & 96.7 & 94.6 & \textbf{97.3} & \textbf{95.3} & 95.9 & 93.3 \\
    010\_potted\_meat\_can & 91.7 & 83.6 & 86.4 & 80.4 & 91.3 & 82.5 & 92.5 & 84.5 & \textbf{94.6} & \textbf{84.9} \\
    011\_banana & 92.6 & 84.6 & \textbf{95.7} & \textbf{90.1} & 95.3 & \textbf{90.1} & 94.7 & 89.4 & 95.3 & 89.8 \\
    019\_pitcher\_base & 96.4 & 93.4 & 95.8 & 91.9 & 96.4 & 93.2 & 96.7 & 93.9 & \textbf{96.8} & \textbf{94.4} \\
    021\_bleach\_cleanser & 89.5 & 80.0 & 90.6 & 83.2 & 90.5 & 82.3 & 90.3 & 82.8 & \textbf{91.5} & \textbf{85.1} \\
    024\_bowl* & 37.1 & 37.1 & 82.5 & 82.5 & 63.9 & 63.9 & 41.8 & 42.5 & \textbf{93.3} & \textbf{93.3} \\
    025\_mug & 96.1 & 90.8 & \textbf{96.9} & \textbf{92.7} & 96.5 & 92.3 & 96.7 & 91.7 & 95.8 & 90.0 \\
    035\_power\_drill & 95.0 & 89.7 & 94.7 & 88.8 & 95.4 & 90.8 & \textbf{95.6} & \textbf{91.4} & 94.8 & 89.7 \\
    036\_wood\_block* & 84.5 & 84.5 & 68.3 & 68.3 & 81.2 & 81.2 & \textbf{87.8} & \textbf{88.1} & 79.6 & 79.6 \\
    037\_scissors & 92.5 & 84.5 & 91.7 & 81.6 & 88.3 & 79.0 & \textbf{93.1} & \textbf{86.5} & \textbf{93.1} & 85.9 \\
    040\_large\_marker & 80.4 & 69.5 & 83.3 & 72.3 & 77.6 & 68.5 & \textbf{85.6} & \textbf{72.5} & 80.3 & 69.4 \\
    051\_large\_clamp* & 85.6 & 85.6 & \textbf{90.0} & \textbf{90.0} & 86.8 & 86.8 & 89.3 & 88.6 & 89.3 & 89.3 \\
    052\_extra\_large\_clamp* & 92.5 & 92.5 & 91.6 & 91.6 & \textbf{94.6} & \textbf{94.6} & 92.7 & 92.7 & 93.9 & 93.9 \\
    061\_foam\_brick* & 95.3 & 95.3 & 94.1 & 94.1 & 93.2 & 93.2 & \textbf{96.5} & 95.7 & \textbf{96.5} & \textbf{96.5} \\ \midrule
    Mean & 90.1 & 85.3 & 91.3 & 86.4 & 90.8 & 86.1 & 91.5 & 87.0 & \textbf{93.1} & \textbf{88.9} \\ \bottomrule
  \end{tabular}
\end{table*}

\section{TUDL Evaluation per Object}
\label{sec:supp_tudl_perobj}
Tables~\ref{tab:app_tudl_add002}, \ref{tab:app_tudl_add005}, and \ref{tab:app_tudl_add01} provide object-level results on TUDL~\cite{hodan2018bop} under ADD 0.02d, 0.05d, and 0.1d, respectively.
GHLAPose achieves the best mean accuracy under all three thresholds, demonstrating consistent improvements in both fine-grained and coarse-grained pose accuracy.
At the object level, our method achieves the best or second-best accuracy on all three objects under each threshold.

\begin{table*}[!t]
  \centering
  \refstepcounter{table}
  \begin{justify}
    \small Table \thetable: Object-level results on TUDL~\cite{hodan2018bop} w.r.t. ADD 0.02d. We highlight the best results in \textbf{bold} and \underline{underline} the second-best results.
  \end{justify}
  \label{tab:app_tudl_add002}
  \setlength{\tabcolsep}{8pt}
  \begin{tabular}{lcccc}
    \toprule
    Method & obj\_000001 & obj\_000002 & obj\_000003 & Mean \\ \midrule
    HccePose~\cite{Wang_2025_ICCV} & \underline{29.5} & 5.5 & 19.0 & 18.0 \\
    ZebraPose~\cite{su2022zebrapose} & 25.5 & \textbf{9.5} & \textbf{23.0} & \underline{19.3} \\
    RACE6D~\cite{Ha_2026_CVPR} & 18.0 & 3.0 & 15.0 & 12.0 \\
    GDRNPP~\cite{gdrnpp} & 15.0 & 5.0 & 16.5 & 12.2 \\
    PFA-CosyPose~\cite{hu2022perspective} & 15.5 & \underline{6.5} & 18.0 & 13.3 \\
    \midrule
    Ours & \textbf{31.5} & \textbf{9.5} & \underline{20.0} & \textbf{20.3} \\
    \bottomrule
  \end{tabular}
\end{table*}

\begin{table*}[!t]
  \centering
  \refstepcounter{table}
  \begin{justify}
    \small Table \thetable: Object-level results on TUDL~\cite{hodan2018bop} w.r.t. ADD 0.05d. We highlight the best results in \textbf{bold} and \underline{underline} the second-best results.
  \end{justify}
  \label{tab:app_tudl_add005}
  \setlength{\tabcolsep}{8pt}
  \begin{tabular}{lcccc}
    \toprule
    Method & obj\_000001 & obj\_000002 & obj\_000003 & Mean \\ \midrule
    HccePose~\cite{Wang_2025_ICCV} & \textbf{68.5} & 24.5 & \underline{49.0} & \underline{47.3} \\
    ZebraPose~\cite{su2022zebrapose} & 55.5 & \textbf{30.5} & 48.5 & 44.8 \\
    RACE6D~\cite{Ha_2026_CVPR} & 50.5 & 22.5 & 45.0 & 39.3 \\
    GDRNPP~\cite{gdrnpp} & 44.5 & 23.5 & 41.0 & 36.3 \\
    PFA-CosyPose~\cite{hu2022perspective} & 42.5 & 22.5 & 41.5 & 35.5 \\
    \midrule
    Ours & \underline{64.0} & \underline{29.5} & \textbf{53.5} & \textbf{49.0} \\
    \bottomrule
  \end{tabular}
\end{table*}

\begin{table*}[!t]
  \centering
  \refstepcounter{table}
  \begin{justify}
    \small Table \thetable: Object-level results on TUDL~\cite{hodan2018bop} w.r.t. ADD 0.1d. We highlight the best results in \textbf{bold} and \underline{underline} the second-best results.
  \end{justify}
  \label{tab:app_tudl_add01}
  \setlength{\tabcolsep}{8pt}
  \begin{tabular}{lcccc}
    \toprule
    Method & obj\_000001 & obj\_000002 & obj\_000003 & Mean \\ \midrule
    HccePose~\cite{Wang_2025_ICCV} & \textbf{92.5} & 47.5 & \underline{80.0} & \underline{73.3} \\
    ZebraPose~\cite{su2022zebrapose} & 84.5 & \underline{51.5} & 73.5 & 69.8 \\
    RACE6D~\cite{Ha_2026_CVPR} & 74.5 & 45.0 & 72.5 & 64.0 \\
    GDRNPP~\cite{gdrnpp} & 73.0 & 47.5 & 68.5 & 63.0 \\
    PFA-CosyPose~\cite{hu2022perspective} & 75.5 & 42.5 & 63.0 & 60.3 \\
    \midrule
    Ours & \underline{92.0} & \textbf{55.0} & \textbf{85.5} & \textbf{77.5} \\
    \bottomrule
  \end{tabular}
\end{table*}

\section{Additional Qualitative Results}
\label{sec:supp_qualitative}
\begin{figure*}[!t]
  \centering
  \refstepcounter{figure}
  \includegraphics[width=\linewidth,height=0.86\textheight,keepaspectratio]{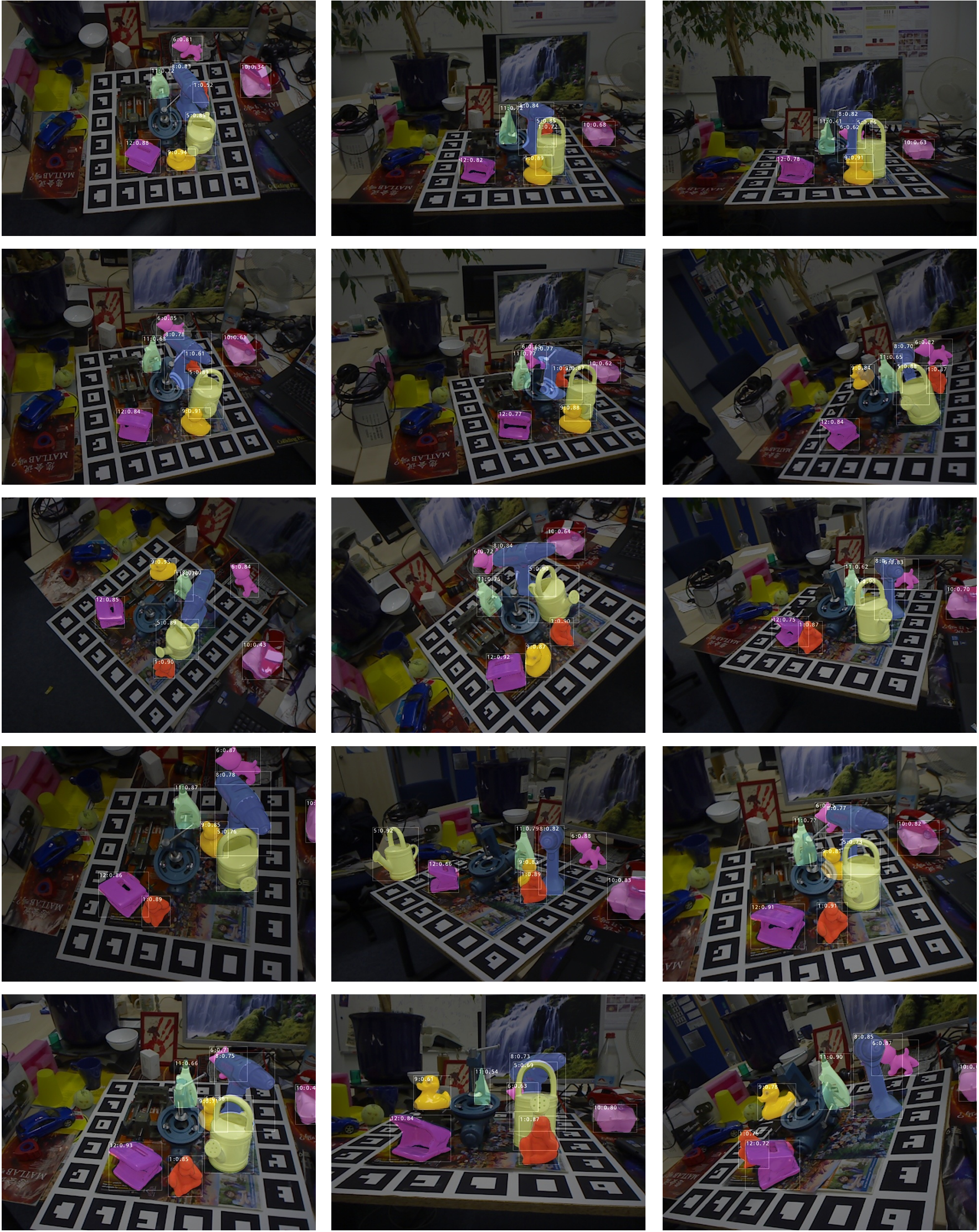}
  \label{fig:supp_pose_vis_lmo}
  \vspace{-8pt}
  \begin{justify}
    \small \textbf{Fig. \thefigure: Qualitative results on LM-O~\cite{LMO}.}
    We render CAD models with estimated poses on top of the original images.
    The confidence scores are from the FCOS detector~\cite{tian2019fcos}.
  \end{justify}
\end{figure*}

\begin{figure*}[!t]
  \centering
  \refstepcounter{figure}
  \includegraphics[width=\linewidth,height=0.86\textheight,keepaspectratio]{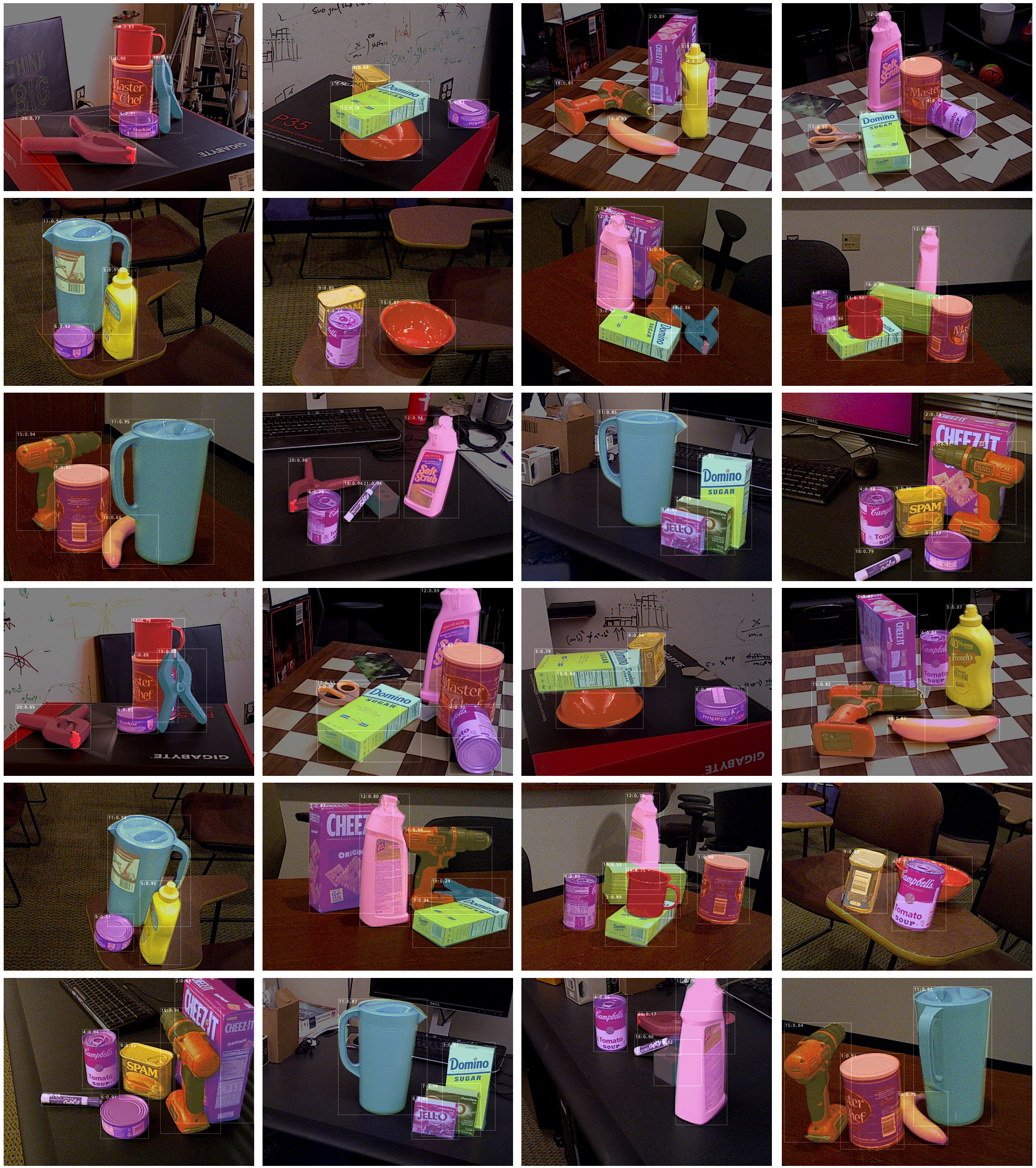}
  \label{fig:supp_pose_vis_ycbv}
  \vspace{-8pt}
  \begin{justify}
    \small \textbf{Fig. \thefigure: Qualitative results on YCB-V~\cite{YCB-V}.}
    We render CAD models with estimated poses on top of the original images.
    The confidence scores are from the FCOS detector~\cite{tian2019fcos}.
  \end{justify}
\end{figure*}

Figs.~\ref{fig:supp_pose_vis_lmo} and~\ref{fig:supp_pose_vis_ycbv} show additional qualitative results on LM-O~\cite{LMO} and YCB-V~\cite{YCB-V}.
The estimated poses are used to render CAD models onto the original images using the BOP toolkit~\cite{hodavn2020bop}.
These examples show that our method produces accurate alignments under occlusion and clutter.

\end{document}